%% file: main.tex
\documentclass[letterpaper]{article} 
\usepackage{aaai2026}  
\usepackage{times}  
\usepackage{helvet}  
\usepackage{courier}  
\usepackage[hyphens]{url}  
\usepackage{graphicx} 
\usepackage{natbib}  
\usepackage{caption} 
\usepackage{algorithm}
\usepackage{algorithmic}

\usepackage{newfloat}
\usepackage{listings}
\DeclareCaptionStyle{ruled}{labelfont=normalfont,labelsep=colon,strut=off} 
\floatstyle{ruled}
\newfloat{listing}{tb}{lst}{}
\floatname{listing}{Listing}
\usepackage{multirow}
\usepackage{booktabs}
\usepackage{makecell}
\usepackage{xcolor}
\usepackage{vcell}
\usepackage{colortbl}
\usepackage{amssymb}
\usepackage{pifont}
\usepackage{adjustbox}
\usepackage{amsmath}
\usepackage{tabularx}
\usepackage[toc]{multitoc}
\usepackage[skins]{tcolorbox}
\usepackage{soul}
\usepackage{subcaption}
\usepackage{fvextra}
\usepackage{enumitem}

\definecolor{wkred}{RGB}{240, 190, 190}
\definecolor{wkblue}{RGB}{190, 210, 235}
\definecolor{wkgreen}{RGB}{190, 225, 200}
\definecolor{wkpurple}{RGB}{210,210,253}
\definecolor{wkyellow}{RGB}{255,241,177}
\definecolor{wkgold}{RGB}{255, 223, 129}
\definecolor{wksilver}{RGB}{192, 192, 192}

\definecolor{codegreen}{rgb}{0,0.6,0}
\definecolor{codegray}{rgb}{0.5,0.5,0.5}
\definecolor{codepurple}{rgb}{0.58,0,0.82}
\definecolor{backcolour}{rgb}{0.95,0.95,0.92}

\definecolor{upred}{HTML}{DC143C}
\definecolor{downgreen}{HTML}{32CD32}

\newcommand{\up}[1]{\textcolor{upred}{{+#1}}}

\newcommand{\best}{\cellcolor{wkred}}
\newcommand{\second}{\cellcolor{wkblue}}

\newcommand{\promptsectionfirst}[1]{%
  \noindent\textbf{\normalsize #1}\par\nopagebreak\vspace{0.5mm}%
}

\newcommand{\promptsectioncompact}[1]{%
  \tcbline*{}%
  \vspace{0.1mm}%
  \noindent\textbf{\normalsize #1}\par\nopagebreak\vspace{0.5mm}%
}

\title{From Solver to Tutor: Evaluating the Pedagogical Intelligence of LLMs with KMP-Bench}

\author{
    Weikang Shi\equalcontrib\textsuperscript{\rm 1},
    Houxing Ren\equalcontrib\textsuperscript{\rm 1},
    Junting Pan\equalcontrib\textsuperscript{\rm 1,\rm 3},
    Aojun Zhou\textsuperscript{\rm 1},
    Ke Wang\textsuperscript{\rm 1},
    Zimu Lu\textsuperscript{\rm 1},\\
    Yunqiao Yang\textsuperscript{\rm 1},
    Yuxuan Hu\textsuperscript{\rm 1},
    Linda Wei\textsuperscript{\rm 1}, 
    Mingjie Zhan\textsuperscript{\rm 1}, 
    Hongsheng Li\thanks{Corresponding author.}\textsuperscript{\rm 1,\rm 2,\rm 3}
}
\affiliations{
    \textsuperscript{\rm 1}Multimedia Laboratory (MMLab), The Chinese University of Hong Kong\\
    \textsuperscript{\rm 2}Shenzhen Loop Area Institute
    \textsuperscript{\rm 3}CPII under InnoHK

    \{wkshi@link, hsli@ee\}.cuhk.edu.hk
}

\begin{document}

\maketitle

\input{latex/00-Abstract}

\input{latex/01-Introduction}

\input{latex/03-Dataset}

\input{latex/04-Evaluation}

\input{latex/05-Experiments}

\input{latex/02-RelatedWork}

\input{latex/06-Conclusion}

\clearpage

\bibliography{aaai2026}

\clearpage

\appendix
\setcounter{secnumdepth}{2} 

\input{latex/Appendix}

\end{document}

%% file: latex/00-Abstract.tex
\begin{abstract}
Large Language Models (LLMs) show significant potential in AI mathematical tutoring, yet current evaluations often rely on simplistic metrics or narrow pedagogical scenarios, failing to assess comprehensive, multi-turn teaching effectiveness. In this paper, we introduce KMP-Bench, a comprehensive K-8 Mathematical Pedagogical Benchmark designed to assess LLMs from two complementary perspectives. The first module, KMP-Dialogue, evaluates holistic pedagogical capabilities against six core principles (e.g., Challenge, Explanation, Feedback), leveraging a novel multi-turn dialogue dataset constructed by weaving together diverse pedagogical components. The second module, KMP-Skills, provides a granular assessment of foundational tutoring abilities, including multi-turn problem-solving, error detection and correction, and problem generation. Our evaluations on KMP-Bench reveal a key disparity: while leading LLMs excel at tasks with verifiable solutions, they struggle with the nuanced application of pedagogical principles. Additionally, we present KMP-Pile, a large-scale (150K) dialogue dataset. Models fine-tuned on KMP-Pile show substantial improvement on KMP-Bench, underscoring the value of pedagogically-rich training data for developing more effective AI math tutors.
\end{abstract}

%% file: latex/01-Introduction.tex
\section{Introduction}

\begin{figure*}[t]
 \centering
 \includegraphics[width=1.0\linewidth]{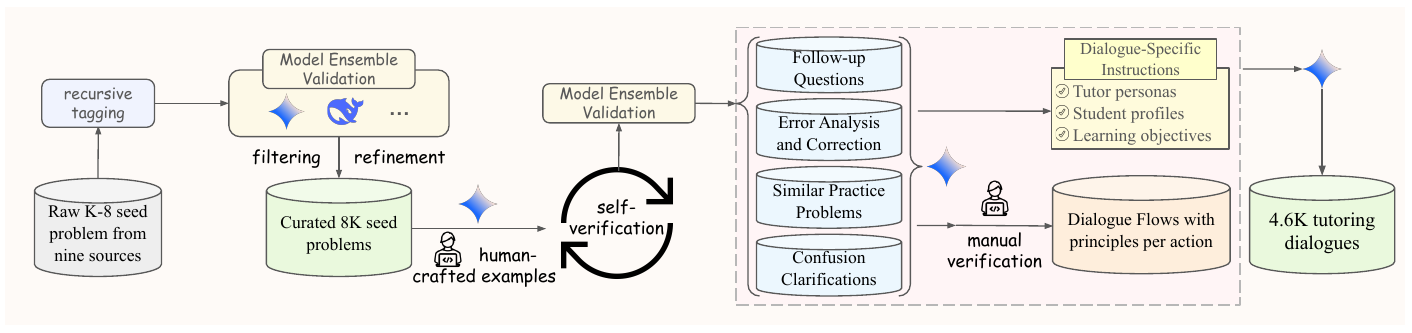}
 \caption{The curation pipeline for the KMP-Bench dataset. The process generates four distinct pedagogical components from curated K-8 problems using an LLM guided by human-crafted few-shot examples. After a rigorous quality control process (including LLM-based self-verification and model ensemble validation), these components are organized into Dialogue Flows, which are then manually verified for pedagogical soundness before being woven into final tutoring dialogues.}
 \label{fig:method_pipeline}
 \vspace{-1mm}
\end{figure*}

The rapid development of Large Language Models~(LLMs) has transformed numerous fields, with education emerging as a particularly promising application area. LLMs have demonstrated considerable potential in AI tutoring, assisting with tasks like essay composition~\citep{shao2024assistingwritingwikipedialikearticles} and providing emotional support~\citep{Shin_2022,kang2025largelanguagemodelsgood}. However, mathematical tutoring presents unique challenges due to its demand for specialized knowledge, distinct pedagogical scenarios, and the difficulty in assessing true teaching effectiveness. While LLMs are increasingly adept at solving mathematical problems \citep{wang2023mathcoderseamlesscodeintegration,yang2024qwen25mathtechnicalreportmathematical, deepseekai2025deepseekr1incentivizingreasoningcapability}, proficiency in problem-solving does not equate to skilled teaching. Expert instruction requires a broader suite of capabilities, including adaptive challenging, effective modeling, guided practice, constructive feedback, etc.~\citep{Allison2015Making}. Consequently, comprehensively evaluating the pedagogical abilities of LLMs for mathematical tutoring presents a significant and pressing challenge.

Current evaluation efforts for LLMs in mathematical tutoring present some limitations. One stream relies on objective metrics like problem-solving accuracy~\citep{Liang2024MathChatBM,Mishra2024MathCAMPSFS} or textual similarity scores (e.g., BLEU, BERTScore)~\citep{ding2024boostinglargelanguagemodels,christ-etal-2024-mathwell}, which, while easily quantifiable, inadequately reflect a model’s teaching efficacy and interactional quality within dynamic, real-world educational settings. Another stream incorporates pedagogical dimensions but mainly uses existing datasets (e.g., MathDial~\citep{Macina2023MathDial}, Bridge~\citep{wang-etal-2024-bridging}) where evaluations are confined to the narrow scenario of error correction. This overlooks a spectrum of critical tutoring functions, such as proactive follow-up questioning, student confusion clarification, and the guided generation of practice problems—all hallmarks of effective, adaptive teaching.
This focus also restricts problem scope (often elementary arithmetic problems) and dataset size.  
This gap means current assessments may not fully reveal how well AI tutors align with effective teaching practices.

To bridge these gaps, we introduce a systematic framework for building and evaluating AI math tutors. The foundation of this framework is a multi-stage data curation pipeline that generates pedagogically-rich, multi-functional tutoring dialogues. 
This pipeline first crafts four distinct pedagogical components—follow-up questions, error analysis scenarios, similar practice problems, and confusion clarifications—and then meticulously weaves them into coherent conversational flows that simulate authentic K-8 tutoring sessions. Applying this pipeline, we create two principal assets: an evaluation set of 4.6K dialogues, and a large-scale (150K) training dataset.

Built upon our evaluation set, we present \textbf{KMP-Bench}, a benchmark suite designed to assess AI math tutors from two complementary perspectives. The first module, \textbf{KMP-Dialogue}, evaluates \textit{holistic pedagogical capabilities}. It assesses a model's ability to generate contextually appropriate responses within truncated dialogues, measuring performance against six core pedagogical principles adapted from established educational guidelines (Challenge, Explanation, Modelling, Practice, Questioning, and Feedback)~\citep{Allison2015Making}.

The second module, \textbf{KMP-Skills}, provides a more granular assessment of \textit{foundational tutoring abilities}. It leverages the individual pedagogical components to create targeted evaluations for three specific skills: (1) multi-turn follow-up problem-solving, (2) error detection and correction, and (3) the generation of valid and pedagogically-sound mathematical problems. Together, KMP-Dialogue and KMP-Skills offer a robust evaluation of an AI tutor's effectiveness across a wide spectrum of K-8 mathematical content.

Our extensive evaluations using KMP-Bench reveal a significant disparity: while current LLMs show promise in structured tasks like problem-solving, they struggle with the nuanced application of pedagogical principles and the reliable generation of pedagogically-sound responses. Furthermore, models fine-tuned on KMP-Pile dataset demonstrate substantial improvements across the KMP-Bench evaluations, underscoring the critical value of pedagogically-rich training data. Our main contributions are:

\begin{enumerate}

\item We introduce a comprehensive benchmark, KMP-Bench, composed of two distinct modules for multi-faceted evaluation: KMP-Dialogue for assessing holistic, principle-based pedagogical capabilities, and KMP-Skills for performing a granular assessment of foundational tutoring abilities.

\item We provide a detailed empirical analysis that reveals a key weakness in modern LLMs: a trade-off between task-solving proficiency and pedagogical quality. Our results highlight the urgent need to focus research on improving pedagogical reasoning and reliable content generation.

\item We release KMP-Pile, a large-scale (150K) training dataset of multi-turn tutoring dialogues. We demonstrate that training on this dataset significantly enhances AI tutor performance, serving as a valuable resource for the community.

\end{enumerate}

%% file: latex/03-Dataset.tex
\begin{figure*}[t]
  \centering
  
  \begin{subfigure}[t]{0.37\textwidth}
    \centering
    \includegraphics[width=\linewidth]{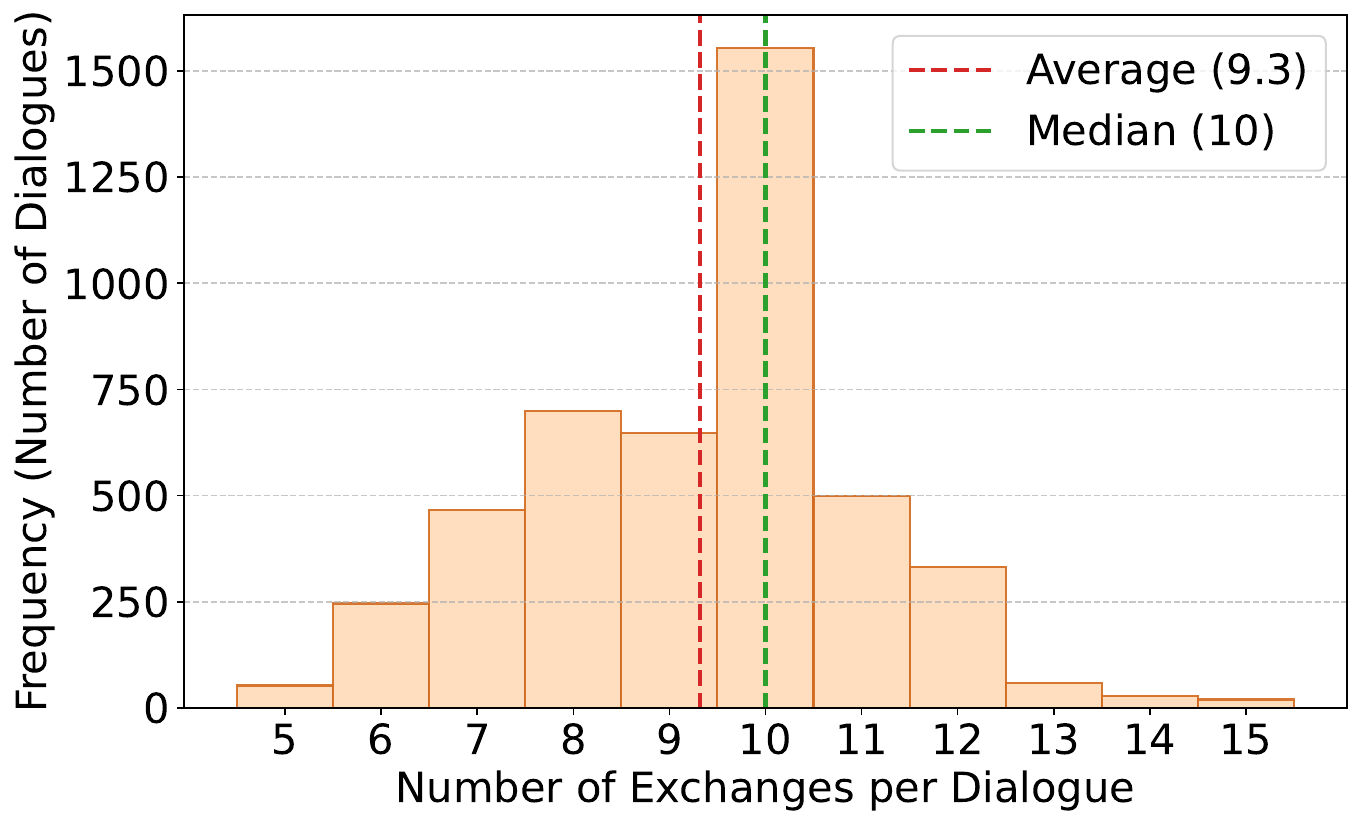}
    \caption{Distribution of Exchanges}
    \label{fig:exchange_dist}
  \end{subfigure}
  \begin{subfigure}[t]{0.37\textwidth}
    \centering
    \includegraphics[width=\linewidth]{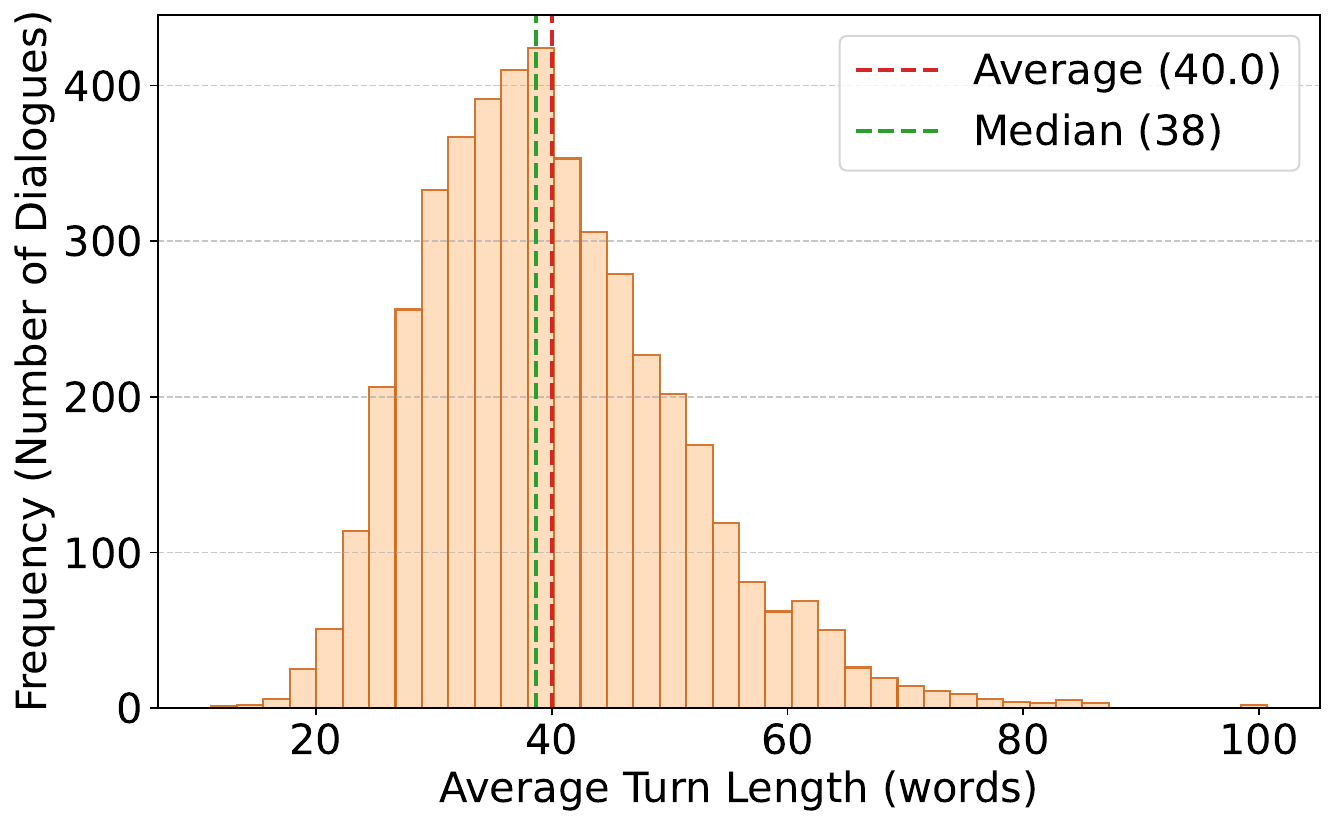}
    \caption{Distribution of Avg. Turn Length}
    \label{fig:avg_len_dist}
  \end{subfigure}
  \begin{subfigure}[t]{0.25\textwidth}
    \centering
    \includegraphics[width=\linewidth]{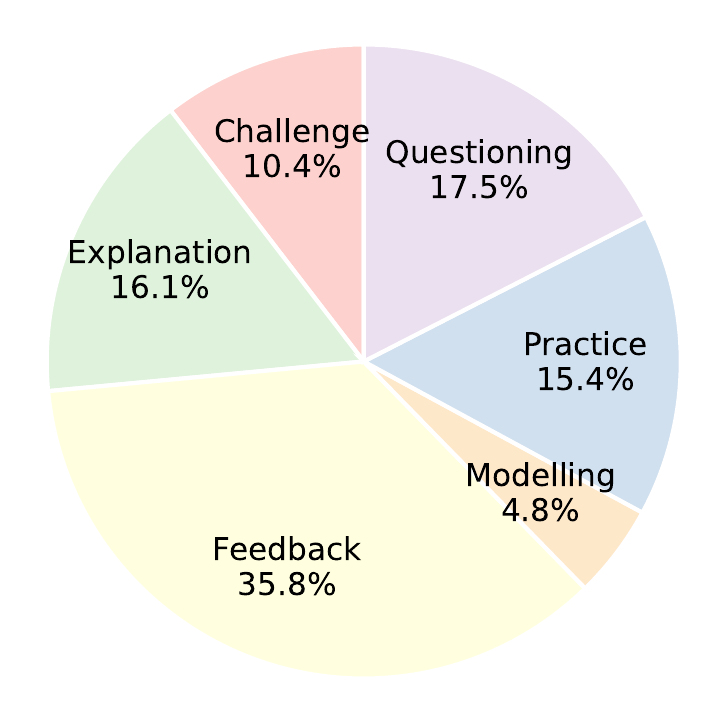}
    \caption{Distribution of Principles}
    \label{fig:principles_dist}
  \end{subfigure} 

  \caption{Statistical distributions of the dataset.}
  \label{fig:all_distributions}
\end{figure*}

\section{Tutoring Dialogue Curation}

This section details our data curation pipeline, illustrated in Figure~\ref{fig:method_pipeline}, which systematically transforms curated K-8 seed problems into a corpus of pedagogically diverse and instructionally sound tutoring dialogues. The process unfolds in three main phases: establishing a seed problem foundation, crafting distinct pedagogical components, and weaving these components into coherent conversational flows.

\subsection{Seed Problem Foundation}

To overcome the limited K-8 curriculum coverage and small scale of existing mathematical educational evaluation datasets like MathDial and Bridge, we established a comprehensive foundation of seed problems. This involved an initial collection from nine diverse elementary to middle school mathematics datasets, followed by a systematic categorization using the recursive tagging algorithm from \citet{lucy2024mathfishevaluatinglanguagemodel} based on Common Core Standards, resulting in problems spanning 11 distinct mathematical domains. To meet the rigorous quality standards required for a benchmark, each problem underwent model ensemble validation for correctness, while its corresponding solution was subsequently refined and standardized. 
This meticulous process, detailed further in Appendix~\ref{app:seed_problem_curation}, yielded 8K validated K-8 mathematical problems that serve as the foundation for our tutoring dialogues.

\subsection{Crafting Pedagogical Interaction Components}
\label{sec:crafting_components}

Building upon the curated seed problems, we systematically generated four types of core conversational components designed to embody distinct pedagogical functions. 
To ensure the quality and relevance of the outputs, each generation task was guided by a set of human-crafted few-shot examples. This foundational step provided the LLM with clear, exemplary precedents for each component type. 
This generation process primarily utilized the Gemini-2.0-Flash model.
Detailed methodologies, including specific prompting strategies and validation protocols, are provided in Appendix~\ref{app:component_generation_details}.

\textbf{Knowledge-Extending Follow-up Questions} are designed to deepen conceptual understanding and foster cognitive extension. 
For each seed problem, we generated a sequence of follow-up questions where each subsequent question was carefully scaffolded to incrementally increase difficulty and explore related mathematical concepts. 
This process included guided reasoning to ensure pedagogical relevance and robust validation of the questions' accuracy.

\textbf{Error Analysis and Correction} scenarios were constructed to develop students' critical thinking and error analysis capabilities. This process began by systematically generating plausible incorrect solutions for the seed problems, by employing varied modeling techniques and parameters designed to elicit common mathematical errors. To ensure a consistent basis for analysis, all solutions, whether correct or incorrect, were then standardized into a uniform step-by-step format, preserving the original reasoning path. Subsequently, an LLM was utilized to generate detailed correction feedback. This feedback involved analyzing the presented solution, pinpointing the first erroneous step if an error existed, and providing a rectified solution from that step forward. A multi-stage verification protocol was also implemented to ensure the quality, consistency, and pedagogical value of these error analysis components.

\textbf{Similar Practice Problems} were generated to reinforce skill acquisition by providing targeted practice across a spectrum of difficulty levels. 
Guided by Common Core Standards annotations, we prompted an LLM to create new problems at three distinct difficulty tiers—easy, medium, and hard—using the original seed problem as the medium-difficulty reference. The generation process incorporated guided reasoning stages for the LLM and subsequent validation, including measures to ensure problem diversity and avoid trivial variations of the seed problem.

\textbf{Confusion Clarifications} dialogues were designed to resolve common points of confusion encountered by K-8 students and provide targeted, explanatory interactions. Using curated examples of genuine student queries, we prompted an LLM to generate multi-turn interactions between a simulated K-8 student and a tutor. These dialogues specifically aim to clarify conceptual understanding (What), elucidate solution steps and methods (How), and explore the underlying rationale for chosen approaches (Why), thereby fostering a deeper comprehension of the mathematical content. 

Upon completing the generation process for all components, we filtered our collection to retain the 6K seed problems for which all four pedagogical component types were successfully created. This curated set forms the foundation for our KMP-Skills evaluation, as detailed in Section~\ref{sec:specific_skills_evaluation}.

\begin{figure*}[t]
 \centering
 \includegraphics[width=1.0\linewidth]{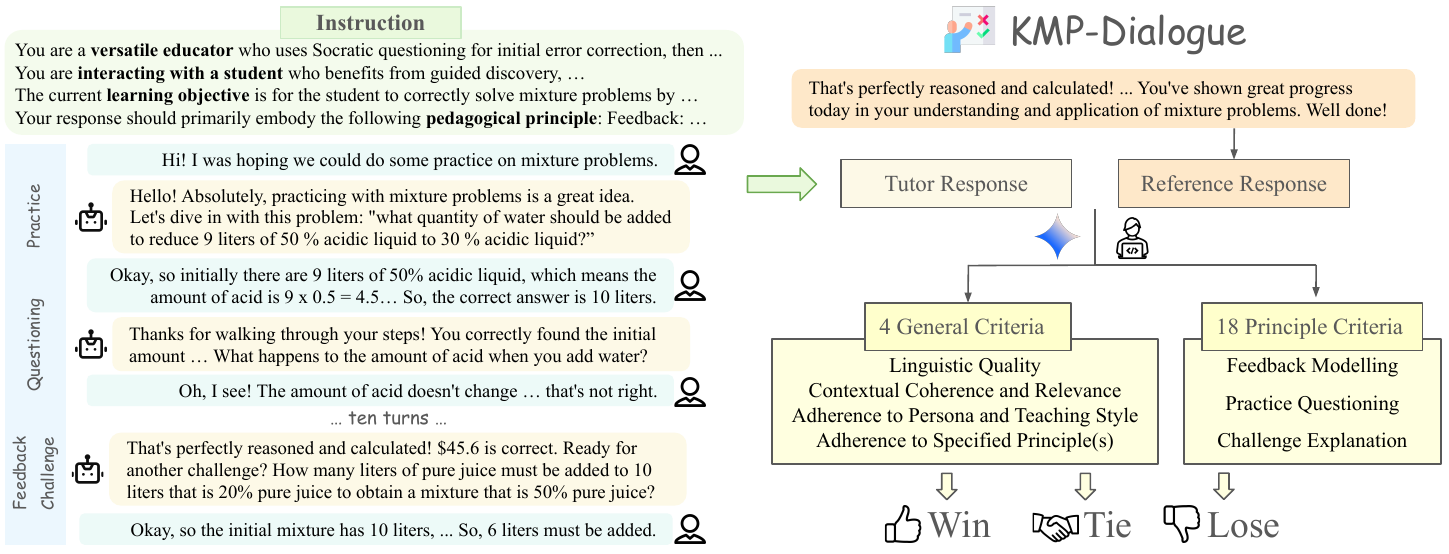}
 \caption{The evaluation framework of KMP-Dialogue. The process evaluates a model's Tutor Response in the context of a full dialogue history and specific instructions (left panel). An LLM or human evaluator compares the Tutor Response against a Reference Response using the 4 general criteria and the criteria corresponding to the relevant pedagogical principle(s) to determine a final ``Win'', ``Tie'', or ``Lose'' outcome.}
 \label{fig:kmp-dialogue}
\end{figure*}

\subsection{Dialogue Weaving and Pedagogical Enrichment}
\label{sec:dialogue_weaving_main} 

To weave the individual pedagogical components into coherent, multi-turn dialogues, we developed a pipeline designed to simulate K-8 tutoring sessions. The process systematically structures the conversation and enriches it with pedagogical diversity and personalized context. Detailed prompts for each stage are provided in Appendix~\ref{app:dialogue_construction_prompts}.

\textbf{Dialogue Flow Generation and Verification.} 
The pipeline begins with a Flow Generator LLM that creates a ``Dialogue Flow''—a structured blueprint for the entire tutoring interaction. This LLM designs a pedagogical ``storyline'' by selecting and sequencing the relevant material components. This stage is where the core pedagogical enrichment occurs: the LLM defines tutor personas, student profiles with potential knowledge gaps, and specific learning objectives. Furthermore, it assigns one or two of six core pedagogical principles to each planned tutor action, creating a foundation for varied and adaptive interactions.

Crucially, these LLM-generated Dialogue Flows then undergo manual verification. Human reviewers assessed each flow against shared quality criteria. They were tasked with identifying and addressing critical flaws, such as the hallucination of materials or illogical pedagogical sequencing, and truncating flows to remove conversational turns that lacked pedagogical value. This process resulted in the removal of 451 (7.6\%) problematic flows. The detailed verification protocol is provided in Appendix~\ref{app:manual_annotation}.

\textbf{Dialogue Expansion.}
Following human approval, the verified Dialogue Flow is passed to a Dialogue Expander LLM, which translates the blueprint into a complete, natural-language dialogue. It expands the planned actions, referenced materials, and persona details from the flow into fluent and K-8 appropriate conversational turns. This two-tiered process, with its human-in-the-loop oversight, enables the creation of dialogues that are both instructionally structured and context-aware.

Using this pipeline, we generated the 4.6K dialogues for our benchmark. To scale up data creation, we then adapted the process by replacing the manual verification step with an LLM-based verifier. Applying this modified pipeline to the Orca-Math~\citep{mitra2024orcamathunlockingpotentialslms} seed problems yielded KMP-Pile, a training dataset of 150K dialogues.

\subsection{Dataset Statistics}

The seed problems for our dataset are sourced from a diverse collection of 9 educational platforms, encompassing 11 distinct mathematical domains and spanning 9 grade levels from Kindergarten to Grade 8 (K-8). A detailed breakdown of this distribution is presented in Figure~\ref{fig:seed_problem_distribution}.

Figure~\ref{fig:all_distributions} highlights the substantial nature of our generated dialogues. The dialogues average a robust 9.3 exchanges (Figure~\ref{fig:all_distributions} (a)), a length sufficient for facilitating meaningful, multi-step pedagogical interactions. Furthermore, the considerable average turn length of 40 words (Figure~\ref{fig:all_distributions} (b)) indicates that the tutor's utterances are detailed and explanatory, which is a hallmark of effective instructional content.

%% file: latex/04-Evaluation.tex
\section{Evaluation Framework}

\subsection{KMP-Dialogue: Evaluating Holistic Pedagogical Capabilities}

\label{sec:pedagogical_evaluation}

To rigorously assess the pedagogical capabilities of tutor LLMs, we developed a multifaceted evaluation framework, as show in Figure~\ref{fig:kmp-dialogue}. 
The framework comprises three main stages: setting up evaluation instances through dialogue truncation, generating dialogue-specific instructions, and curating a nuanced set of evaluation criteria. 

\textbf{Evaluation Instance Preparation via Dialogue Truncation.}
First, to prepare evaluation instances, we adapt our coherent tutoring interaction dataset. Dialogues are strategically truncated at a tutor's turn, preserving the conversational history up to the preceding student utterance. This creates a test case where a tutor LLM is prompted to generate the subsequent pedagogical response. Each selected tutor turn is guided by one or two of our six core pedagogical principles (Challenge, Explanation, Modelling, Practice, Questioning, and Feedback). Figure~\ref{fig:all_distributions} (c) shows the precise distribution of these principles across the truncated turns used in our evaluation set. Furthermore, an LLM assists in filtering these truncated dialogues, retaining only those deemed to possess significant value for evaluation.

\textbf{Dialogue-Specific Instructions Generation. }
Second, recognizing that effective tutoring responses are highly dependent on context, we generate dialogue-specific instructions for the tutor LLM under evaluation. Since each truncated dialogue originates from a unique scenario with specific tutor personas, student profiles, and learning objectives, this contextual information is vital. We employ an LLM to craft a concise set of instructions for each test case, serving as a system prompt for the tutor LLM. Crucially, these instructions not only convey the necessary scenario details but also explicitly specify the target pedagogical principle(s) that the tutor LLM's response should embody for that particular turn. This approach facilitates targeted, principle-based evaluation and assesses the model's capacity to adhere to specific pedagogical directives. 

\textbf{Context-Aware Evaluation Criteria and Method.}
Finally, to evaluate the generated tutor responses, we curated a comprehensive set of 22 evaluation criteria, moving beyond generic metrics. This involved both manual expert specification and LLM assistance. Our evaluation framework is distinguished by its two-tiered structure—comprising four general criteria applicable to all responses and three principle-specific criteria for each of the six pedagogical principles—and its dialogue-specific nature. The criteria are adapted to consider the unique context of each interaction, such as the established tutor persona or student needs, enhancing evaluation validity. For assessment, the tutor LLM's generated response is compared against the original tutor turn (which was truncated from the dialogue and serves as a reference response). 
This comparison yields a Win, Tie, or Lose outcome for each applicable criterion and an overall judgment for the tutor's entire response. 
This multi-layered feedback enables a nuanced analysis of the tutor LLM's pedagogical strengths and weaknesses in multi-turn conversational settings. 
The grading prompt is available in Appendix~\ref{app:pedagogical_evaluation}.

\subsection{KMP-Skills: Evaluating Foundational Tutoring Abilities}
\label{sec:specific_skills_evaluation}

While the evaluation of our math tutoring dialogues provides insight into a model's overall pedagogical and instructional capabilities, the proficiency in core, granular tutoring skills is also critical. Abilities such as multi-turn problem solving, accurate error detection and correction, and the generation of valid practice problems form the bedrock of effective teaching. 
We use the 6K Pedagogical Interaction Components from Section~\ref{sec:crafting_components} to construct targeted evaluations for these three skills. Collectively, we refer to this suite of evaluations as KMP-Skills.

\textbf{Multi-turn Follow-up Problem-Solving.}
This evaluation assesses a model's continuous reasoning and its ability to handle increasing problem complexity within a dialogue. Adopting a multi-turn conversational structure similar to MathChat~\citep{Liang2024MathChatBM}, we use sequences comprising a seed problem and two connected, progressively harder follow-up questions. The evaluation proceeds by making the model's response to each question the context for the subsequent one.
We calculate solution accuracy independently for each of the three turns. This per-turn metric reveals how a model's performance changes as conversational depth and problem difficulty increase.

\textbf{Error Detection and Correction.}
An effective tutor must identify and correct student errors. This evaluation uses our set of generated student solutions, which includes both correct examples and incorrect ones containing annotated error steps and reasons. 
Inspired by ProcessBench~\citep{zheng2024processbenchidentifyingprocesserrors}, we evaluate the model's core ability to find and fix mistakes. In this setup, the model performs a two-turn task: (1) it identifies whether a solution is correct and, if not, pinpoints the first erroneous step; (2) it provides a fully corrected solution. Performance is measured by two fundamental metrics: the F1-score for the error identification in Turn 1 and the correction accuracy of the solution in Turn 2.

Further, to gain a more nuanced understanding of a model's ability to diagnose not just where an error is but why it occurred, we adopt the MR-Score from the MR-GSM8K framework~\citep{zeng2024mrgsm8kmetareasoningbenchmarklarge}. This composite score offers deeper diagnostic insight by jointly evaluating three model outputs: the final corrected solution, the identified first error step, and the validity of the stated error reason.

\begin{figure}[t]
 \centering
 \includegraphics[width=1.0\columnwidth]{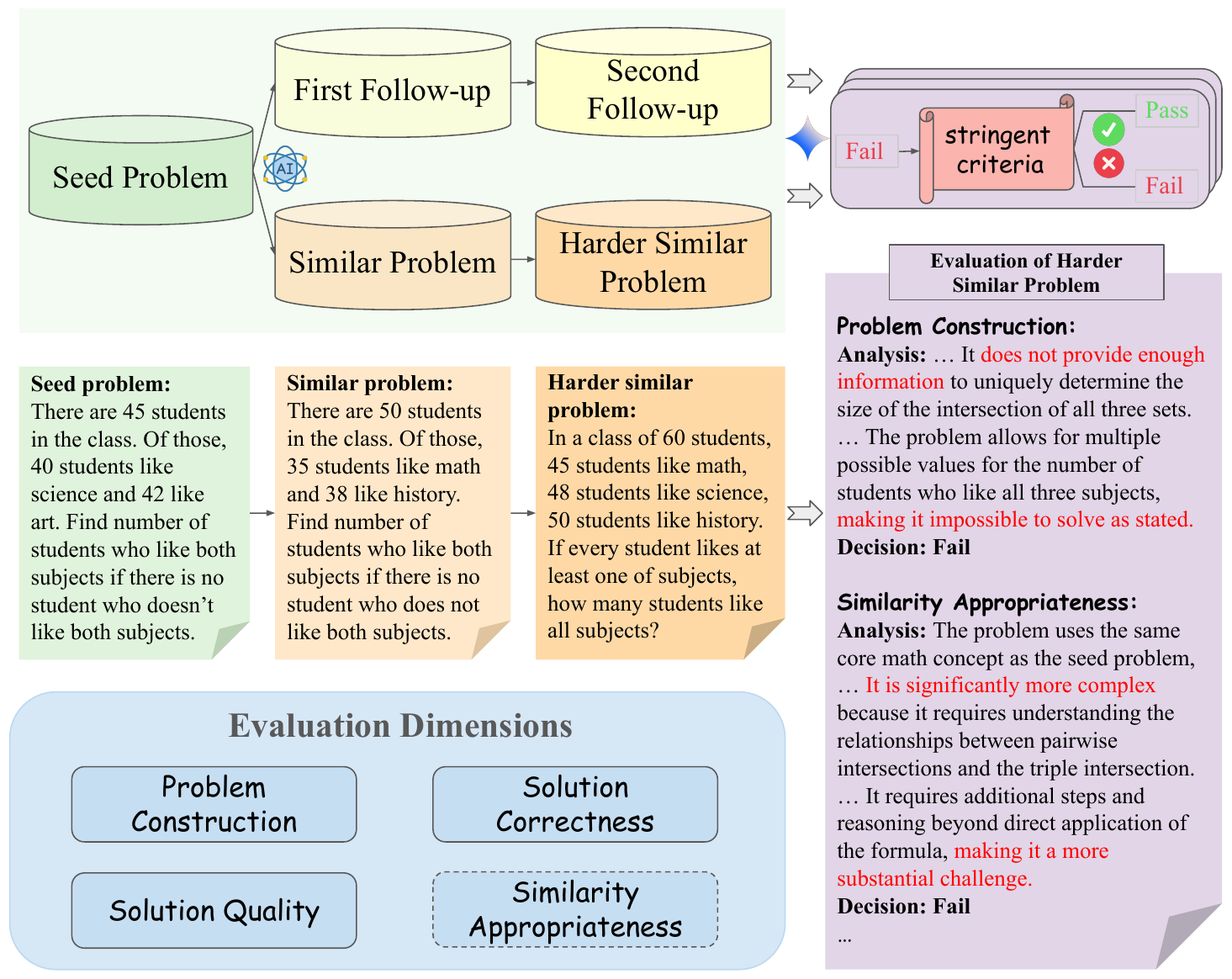}
 \caption{The evaluation framework for the Mathematical Problem Generation task in KMP-Skills.}
 \label{fig:kmp-gen}
\end{figure}

\textbf{Mathematical Problem Generation.}
The ability to generate valid and contextually relevant mathematical problems is crucial for adaptive tutoring systems.
However, ensuring that auto-generated problems and their solutions are logically sound and error-free remains a significant challenge. This evaluation, therefore, assesses a model's ability to generate two key types of problems: \textit{follow-up problems}, which extend concepts from a seed problem with increasing complexity, and \textit{similar problems}, which target the same skills using different contexts or values for reinforcement.

This generation task (illustrated in Figure~\ref{fig:kmp-gen}) employs a three-turn dialogue. Given a seed problem and solution, models sequentially generate two new problems (either follow-up or similar types) along with their solutions, maintaining conversational context. These generated problem-solution pairs are then evaluated by a strong LLM using a strict ``fail-by-default'' binary classification across key dimensions: Problem Construction, Solution Correctness, Solution Quality, and (for similar problems) Similarity Appropriateness. Performance is measured as the average proportion of ``Pass'' judgments across all evaluated dimensions.

\input{tables/performance_KMP-Dialogue}

%% file: tables/performance_KMP-Dialogue.tex
\begin{table*}[ht]
  \centering
  \setlength{\tabcolsep}{1mm}
  \small
  \begin{tabular}{l|c|c|cccccc|c}
    \toprule
    \multirow{2.5}{*}{\textbf{Model}} &
    \multirow{2.5}{*}{\textbf{\parbox{1.8cm}{\centering Overall\\Judgement}}} &
    \multirow{2.5}{*}{\textbf{\parbox{1.8cm}{\centering General-\\Level Acc}}} &
    \multicolumn{6}{c|}{\textbf{Principle-Level Acc}} &
    \multirow{2.5}{*}{\textbf{\parbox{1.8cm}{\centering Overall\\Acc}}} \\
    \cmidrule(lr){4-9}
    ~ & ~ & ~ &
    \textbf{\scriptsize Challenge} & \textbf{\scriptsize Explanation} & \textbf{\scriptsize Modelling} & \textbf{\scriptsize Practice} & \textbf{\scriptsize Questioning} & \textbf{\scriptsize Feedback} &
    ~ \\
    \noalign{\smallskip}
    \toprule
                             \multicolumn{10}{c}{Closed-Source Models}                               \\
    \midrule
GPT-4o & 60.7 & 44.5 & 43.4 & 62.2 & 57.5 & 47.5 & 59.9 & 41.3 & 48.2 \\
    Gemini-2.0-Flash & 58.5 & 38.0 & 51.1 & 49.9 & 34.2 & 42.2 & 50.8 & 37.6 & 41.1 \\
    Claude-3.7-Sonnet & \best{87.1} & \second{69.8} & \best{80.0} & \second{79.3} & 71.2 & \second{69.1} & 79.3 & \best{72.2} & \second{72.5} \\
LearnLM-2.0-Flash & 74.8 & 58.9 & \second{77.3} & 42.7 & 24.9 & 67.6 & 73.5 & 63.9 & 58.6 \\
    \midrule
                             \multicolumn{10}{c}{Open-Source Models}                                 \\
    \midrule

    DeepSeek-V3 & \second{84.6} & \best{71.6} & 70.3 & \best{81.5} & \second{71.4} & \best{69.6} & \best{83.9} & \second{70.8} & \best{73.1} \\
    Qwen2.5-Math-72B-Inst & 31.0 & 27.1 & 25.6 & 51.3 & 42.9 & 33.9 & 29.5 & 23.7 & 30.8 \\
    Qwen2.5-72B-Inst & 68.5 & 55.5 & 50.5 & 77.3 & \best{71.5} & 60.7 & 54.0 & 52.4 & 58.3 \\
Qwen2.5-7B-Inst & 62.5 & 52.9 & 41.2 & 74.2 & 66.5 & 56.6 & 49.7 & 50.6 & 54.7 \\
    LLaMa-3.1-8B & 51.3 & 40.3 & 33.0 & 50.1 & 29.0 & 38.3 & 55.0 & 46.2 & 41.1 \\
    SocraticLM & 21.2 & 15.4 & 17.0 & 25.0 & 21.0 & 21.4 & 27.3 & 17.8 & 18.5 \\
    $\textup{MathChat}_{\textup{sync}}$-LLM & 16.1 & 12.0 & 16.0 & 23.0 & 15.4 & 17.6 & 23.1 & 16.3 & 15.3 \\
    TutorChat-LLM & 32.1 & 30.0 & 21.1 & 41.3 & 33.2 & 37.1 & 40.6 & 29.9 & 31.9 \\
    \midrule
Qwen2.5-Math-7B-Inst & 21.0 & 19.7 & 17.5 & 41.7 & 34.5 & 23.2 & 25.5 & 22.6 & 23.6 \\
\textbf{KMP-LM-7B} & 40.4 & 35.1 & 29.3 & 47.0 & 38.5 & 39.8 & 43.0 & 36.1 & 37.0 \\
$\Delta$ & \up{19.4} & \up{15.4} & \up{11.8} & \up{5.3} & \up{4.0} & \up{16.6} & \up{17.5} & \up{13.5} & \up{13.4} \\
    \bottomrule
  \end{tabular}
  \caption{Overall pedagogical performance of various LLMs in KMP-Dialogue. Tutor LLM responses are compared against reference responses from our curated dialogues. The best and second-best results among all the models are highlighted in red and blue, respectively. The $\Delta$ row represents the improvement of KMP-LM-7B over Qwen2.5-Math-7B-Instruct.} 
  \label{tab:kmpbench-p_performance}
\end{table*}

%% file: latex/05-Experiments.tex
\input{tables/performance_KMP-Skills}

\input{tables/human_grade}

\section{Experiments}
\label{sec:experiments}

\subsection{Experimental Setup}
\label{sec:experimental_setup}

\textbf{Metrics.}
We use Gemini-2.0-Flash as the LLM evaluator. For KMP-Dialogue, all accuracies are win rates. Key metrics include: Overall Judgement Acc. (from the evaluator's holistic Win/Tie/Lose decision); General-Level Acc. (average win rate across 4 general criteria); six Principle-Level Acc. scores (average win rate of 3 specific criteria for each of the 6 pedagogical principles); and a composite Overall Acc. (an average of General-Level Acc. and the mean of the six Principle-Level Accs.). For KMP-Skills, we report only the final-round performance of follow-up problem-solving and problem generation for brevity.

\textbf{Models.}
Our evaluation spans a diverse set of LLMs. This includes: (a) closed-source models: GPT-4o~\citep{openai2024gpt4ocard}, Gemini-2.0-Flash~\citep{google2025gemini}, Claude-3.7-Sonnet~\citep{anthropic2025claude}, and LearnLM-2.0-Flash~\citep{learnlmteam2024learnlmimprovinggeminilearning}; (b) a broad range of open-source general-purpose and math-specialized models such as the LLaMA series, Qwen2.5 series, and DeepSeek-V3~\citep{deepseekai2025deepseekv3technicalreport}; and (c) models with a tutoring focus. Among these are established tutor-specific models like SocraticLM~\citep{NEURIPS2024_9bae399d}. Furthermore, we fine-tuned Qwen2.5-Math-7B on two tutoring datasets, $\textup{MathChat}_{\textup{sync}}$~\citep{Liang2024MathChatBM} and TutorChat~\citep{chevalier2024language}, resulting in $\textup{MathChat}_{\textup{sync}}$-LLM and TutorChat-LLM.

\textbf{Training Details.} 
We developed our primary model, KMP-LM-7B, by fine-tuning Qwen2.5-Math-7B base model. The training data consisted of 150K KMP-Pile directly mixed with 208K general instruction samples, which were created by combining 100K randomly sampled conversations from UltraChat~\citep{ding2023enhancingchatlanguagemodels} with 108K multi-turned WildChat~\citep{zhao2024wildchat1mchatgptinteraction} data. 
More training details, including hyperparameter selection, are presented in Appendix~\ref{more_train_details}.
The $\textup{MathChat}_{\textup{sync}}$-LLM and TutorChat-LLM were fine-tuned with the same number of steps and comparable training configurations.

\subsection{Main Results}
In this section, we present our main evaluation results. Comprehensive results are available in Appendix~\ref{app:more_results}.

\textbf{Overall Pedagogical Performance.}
The overall pedagogical performance of various LLMs on KMP-Dialogue is presented in Table~\ref{tab:kmpbench-p_performance}. Among prominent closed-source models, Claude-3.7-Sonnet achieves the leading Overall Accuracy at 72.5. In the open-source domain, DeepSeek-V3 stands out with the highest Overall Accuracy (73.1), demonstrating that top-tier open-source models are highly competitive. Beyond the leaders, the Qwen2.5 model family offers a particularly insightful finding: the general-purpose Qwen2.5-72B-Inst comprehensively outperforms its math-tuned counterpart, Qwen2.5-Math-72B-Inst, suggesting that while math-specific models are optimized for solving problems, this specialization does not readily translate to the nuanced skill of pedagogical dialogue. 
The principle-level analysis further reveals diverse strengths across models: DeepSeek-V3 excels in Explanation (81.5) and Questioning (83.9), while Claude-3.7-Sonnet is dominant in Challenge (80.0) and Feedback (72.2).

Our fine-tuned model, KMP-LM-7B, demonstrates the effectiveness of our approach, achieving an Overall Accuracy of 37.0. This marks a substantial improvement over Qwen2.5-Math-7B-Inst model, with absolute gains of +13.4 in Overall Accuracy and +19.4 in the holistic Overall Judgement score. This performance also positions KMP-LM-7B favorably against other open-source tutoring-focused models, surpassing SocraticLM (18.5), $\textup{MathChat}_{\textup{sync}}$-LLM (15.3), and TutorChat-LLM (31.9). These results underscore that fine-tuning on the pedagogically structured dialogues within our KMP-Pile dataset is a highly effective strategy for boosting the pedagogically capabilities of LLMs, validating our data curation methodology.

\textbf{Foundational Tutoring Skills Evaluation.}
The performance of various models on KMP-Skills is detailed in Table~\ref{kmpbench-t_performance}. In general, leading closed-source models like Gemini-2.5-Flash show high proficiency in the more structured tasks, achieving a score of 91.2 in Multi-Turn Follow-up Problem-Solving and a 96.7 correction accuracy in Error Detection and Correction. Our fine-tuned model, KMP-LM-7B, achieves a competitive 77.5 in problem-solving and demonstrates substantial improvements over Qwen2.5-Math-7B-Inst model across the board. The gains are most pronounced in Error Detection and Correction (a remarkable +45.6 increase in MR-Score) and Problem Generation (Follow-up: +32.5, Similar: +30.4).

Taken together, our results reveal a critical insight into current AI tutors. While LLMs have become exceptionally proficient at tasks with verifiable solutions, such as problem-solving and error correction, a significant performance gap remains in tasks requiring deep pedagogical awareness. This gap is evident in both the inconsistent performance on teaching principles in KMP-Dialogue and the broader challenge of creating pedagogically sound questions in the Problem Generation task. This distinction underscores that the frontier for AI tutors is shifting from being accurate problem-solvers to becoming effective, student-centric educators.

\subsection{Ablation Studies}

\textbf{Analysis of the Accuracy of KMP-Dialogue Evaluator Results. }
To validate our automated evaluation pipeline, we conducted a human evaluation to assess the reliability of our LLM evaluator, Gemini-2.0-Flash. We randomly selected 300 instances from KMP-Dialogue, covering responses from three top-performing models, and had them annotated by human experts to establish a ground truth. As detailed in Table~\ref{tab:human_eval_alignment}, the results from Gemini-2.0-Flash closely mirror the manual evaluations, with an average Alignment Rate of 89.8\% between the LLM's judgments and the human ground truth. This high level of agreement confirms that our automated evaluation is effective and that the LLM evaluator serves as a reliable proxy for human judgment. Further details on the human evaluation protocol are provided in Appendix~\ref{app:manual_annotation}.

\begin{figure}[t]
 \centering
 \includegraphics[width=1.0\columnwidth]{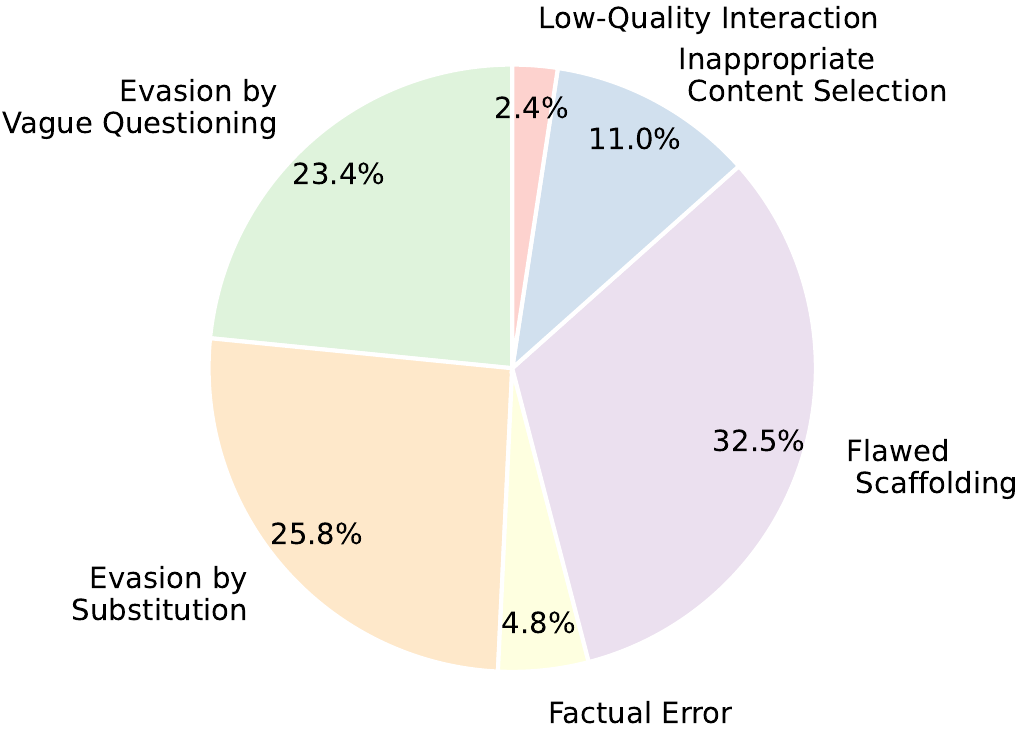}
 \caption{The distribution of the task errors.}
 \label{fig:error_distribution_pie}
 \vspace{-1mm}
\end{figure}

\textbf{Analysis of the Errors in KMP-Dialogue Tasks.}
We analyze the errors and pedagogical flaws that occur in KMP-Dialogue evaluation using mainstream LLMs such as DeepSeek-V3, as illustrated in Figure~\ref{fig:error_distribution_pie}. A detailed explanation of the error types and representative examples can be found in Appendix~\ref{app:error_analysis}. Currently, the most prevalent issues are flawed scaffolding (32.5\%) and evasion by substitution (25.8\%), indicating that LLMs struggle with providing appropriate pedagogical guidance and tend to avoid directly addressing student needs by offering tangential responses.

%% file: tables/performance_KMP-Skills.tex
\begin{table*}[t]
\centering
\small
\begin{tabular}{l|c|c|ccc|cc} 

\noalign{\smallskip}
\toprule 
\multirow{2.5}{*}{\textbf{Model}} & \multirow{2.5}{*}{\textbf{Size}} & \multirow{2.5}{*}{\parbox{1.2cm}{\centering \textbf{KMP-}\\\textbf{FQA}}} & \multicolumn{3}{c|}{\textbf{KMP-EC}} & \multicolumn{2}{c}{\textbf{KMP-Gen}} \\ 

\cmidrule(lr){4-6} \cmidrule(lr){7-8} 
~ & ~ & ~ & \textbf{MR-Score} & \textbf{F1-Score} & \textbf{Correction} & \textbf{Follow-up} & \textbf{Similar} \\

\midrule \multicolumn{8}{c}{Closed-Source Models} \\ \midrule 

GPT-4o & - & 84.1 & 64.4 & 78.3 & 90.9 & 74.8 & 72.2 \\
GPT‑4o mini & - & 79.4 & 56.9 & 66.1 & 84.3 & 76.5 & 64.8 \\
Gemini 2.0 Flash & - & \best{91.2} & \best{86.3} & \best{87.5} & \best{96.7} & 75.0 & 73.2 \\
LearnLM-1.5-Pro & - & \second{87.7} & \second{69.8} & 74.6 & \second{94.4} & 77.0 & 75.0 \\

\midrule \multicolumn{8}{c}{Open-Source Models} \\ \midrule 

Qwen2.5-72B-Inst & 72B & 82.2 & 65.8 & 78.6 & 89.5 & \second{81.3} & 78.1 \\
Qwen2.5-Math-72B-Inst & 72B & 83.2 & 53.8 & 73.5 & 81.0 & \best{86.3} & \second{82.6} \\
Qwen2.5-7B-Inst & 7B & 71.0 & 34.9 & 44.3 & 73.2 & 66.2 & 66.2 \\
Llama-3.1-8B-Inst & 8B & 50.1 & 28.6 & 9.0 & 55.9 & 41.7 & 38.5 \\
$\textup{MathChat}_{\textup{sync}}$-LLM & 7B & 53.5 & 23.5 & 9.8 & 18.9 & 70.9 & 71.3 \\
TutorChat-LLM & 7B & 57.7 & 24.0 & 38.0 & 48.5 & 65.8 & 69.5 \\

\midrule

Qwen2.5-Math-7B-Inst & 7B & 76.1 & 3.4 & 44.1 & 52.3 & 48.6 & 52.8 \\
\textbf{KMP-LM-7B} & 7B & 77.5 & 49.0 & 57.5 & 63.4 & 81.1 & \best{83.2} \\ 
$\Delta$ &  & \up{1.4} & \up{45.6} & \up{13.4} & \up{11.1} & \up{32.5} & \up{30.4} \\
\midrule

\end{tabular}
\caption{Foundational tutoring performance of various LLMs on KMP-Skills. Abbreviations: KMP-FQA denotes Multi-Turn Problem Solving, KMP-EC denotes Error Detection and Correction, and KMP-Gen denotes Problem Generation. The best and second-best results among all the models are highlighted in red and blue, respectively. The $\Delta$ row represents the improvement of KMP-LM-7B over Qwen2.5-Math-7B-Instruct.}
\label{kmpbench-t_performance}
\end{table*}

%% file: tables/human_grade.tex
\begin{table*}[ht]
\centering
\small
\begin{tabular}{llcccc}
\toprule
\textbf{Model} & \textbf{Grading Method} & \textbf{General-Level Acc} & \textbf{Principle-Level Acc} & \textbf{Overall Acc} & \textbf{Alignment Rate (\%)} \\
\midrule
\multirow{2}{*}{LearnLM-2.0-Flash} & Gemini-2.0-Flash & 61.8 & 60.4 & 61.1 & \multirow{2}{*}{89.7} \\
                        & Manual          & 62.5 & 57.9 & 60.2 & \\ 
\midrule
\multirow{2}{*}{Claude-3.7-Sonnet} & Gemini-2.0-Flash & 72.5 & 75.7 & 74.1 & \multirow{2}{*}{87.1} \\
                                   & Manual          & 72.9 & 80.5 & 76.7 & \\ 
\midrule
\multirow{2}{*}{DeepSeek-V3}       & Gemini-2.0-Flash & 75.0 & 75.5 & 75.2 & \multirow{2}{*}{92.5} \\
                                   & Manual          & 75.8 & 76.5 & 76.1 & \\ 
\bottomrule
\end{tabular}
\caption{Comparison of evaluation results between our LLM evaluator (Gemini-2.0-Flash) and manual human annotation on a subset of KMP-Dialogue. The Alignment Rate measures the percentage of criteria where the LLM evaluator's judgment aligns with the human ground truth.}
\label{tab:human_eval_alignment}
\end{table*}

%% file: latex/02-RelatedWork.tex
\section{Related Works}
\label{sec:related_works}

\textbf{Evaluation of AI-based Math Tutors.}
Evaluating AI math tutors requires frameworks that assess the interactive pedagogical process, beyond single-turn problem-solving accuracy. However, many current evaluations still rely on isolated accuracy metrics for discrete skills like multi-turn reasoning~\citep{zheng2023judgingllmasajudgemtbenchchatbot,xu2024supercluemath6gradedmultistepmath,shi2025mathcanvasintrinsicvisualchainofthought,Mishra2024MathCAMPSFS}, error correction~\citep{zheng2024processbenchidentifyingprocesserrors,zeng2024mrgsm8kmetareasoningbenchmarklarge} or problem generation~\citep{Liang2024MathChatBM,christ-etal-2024-mathwell}, failing to capture the quality of the holistic interaction. While datasets like MathDial~\citep{Macina2023MathDial} construct tutoring dialogues for error correction scenarios, their reliance on textual similarity scores (BLEU, BERTScore) correlates poorly with pedagogical quality. 
Furthermore, the problem scope of existing benchmarks is often confined to arithmetic word problems (GSM8K~\citep{cobbe2021trainingverifierssolvemath}). Even specialized benchmarks like MathTutorBench \citep{macina2025mathtutorbenchbenchmarkmeasuringopenended} and MRBench \citep{maurya2025unifyingaitutorevaluation}, while evaluating pedagogical dimensions, tend to assess specific tutoring aspects in isolation.
KMP-Bench addresses these gaps by providing a holistic evaluation framework. It is designed to assess a model's ability to seamlessly integrate key tutoring functions—problem-solving guidance, error correction, and similar problem generation—across a diverse K-8 curriculum, using pedagogically-grounded criteria rather than simplistic metrics.

\textbf{LLMs for Math Education.}
Recent advances in LLMs have spurred their application in building intelligent tutoring systems (ITS) with high pedagogical value~\citep{10050083,Letourneau2025Systematic,computers13100270}. Two primary approaches have emerged. The first involves creating agentic systems~\citep{chowdhury2024autotutormeetslargelanguage,bidarian2023khan,wang2025tutorcopilothumanaiapproach}, which leverage powerful foundation models like GPT-4 \citep{openai2024gpt4technicalreport}. While these systems offer immediate, personalized support, they can lack deep pedagogical grounding and struggle with real-world alignment.
The second approach involves fine-tuning LLMs on tutoring datasets to instill specific capabilities. These datasets are either human-annotated, like MathDial~\citep{Macina2023MathDial}, which are pedagogically authentic but costly to scale, or synthetically generated~\citep{NEURIPS2024_9bae399d,chevalier2024language}. Synthetic approaches offer scalability but often suffer from factual hallucinations and a lack of pedagogical diversity. Our work addresses these limitations by introducing a pipeline that generates dialogues integrating multiple core tutoring functions, ensuring both pedagogical richness and factual accuracy through a structured generation and verification process.

%% file: latex/06-Conclusion.tex
\section{Conclusion}
\label{sec:conclusion}

This paper introduced KMP-Bench, a comprehensive K-8 benchmark with two modules, KMP-Dialogue and KMP-Skills, to assess both holistic, principle-based pedagogical capabilities and foundational tutoring skills. Our evaluations highlight a critical weakness: LLMs are proficient at tasks with verifiable solutions like problem-solving but require significant improvement in applying pedagogical principles and generating valid mathematical problems. Training on our pedagogically-rich dataset, KMP-Pile, improved performance on KMP-Bench, underscoring that such data is crucial for advancing AI from mere solvers to effective tutors.

%% file: latex/Appendix.tex
\label{sec:appendix}

\begin{figure}[t]
 \centering
 \includegraphics[width=1.0\columnwidth]{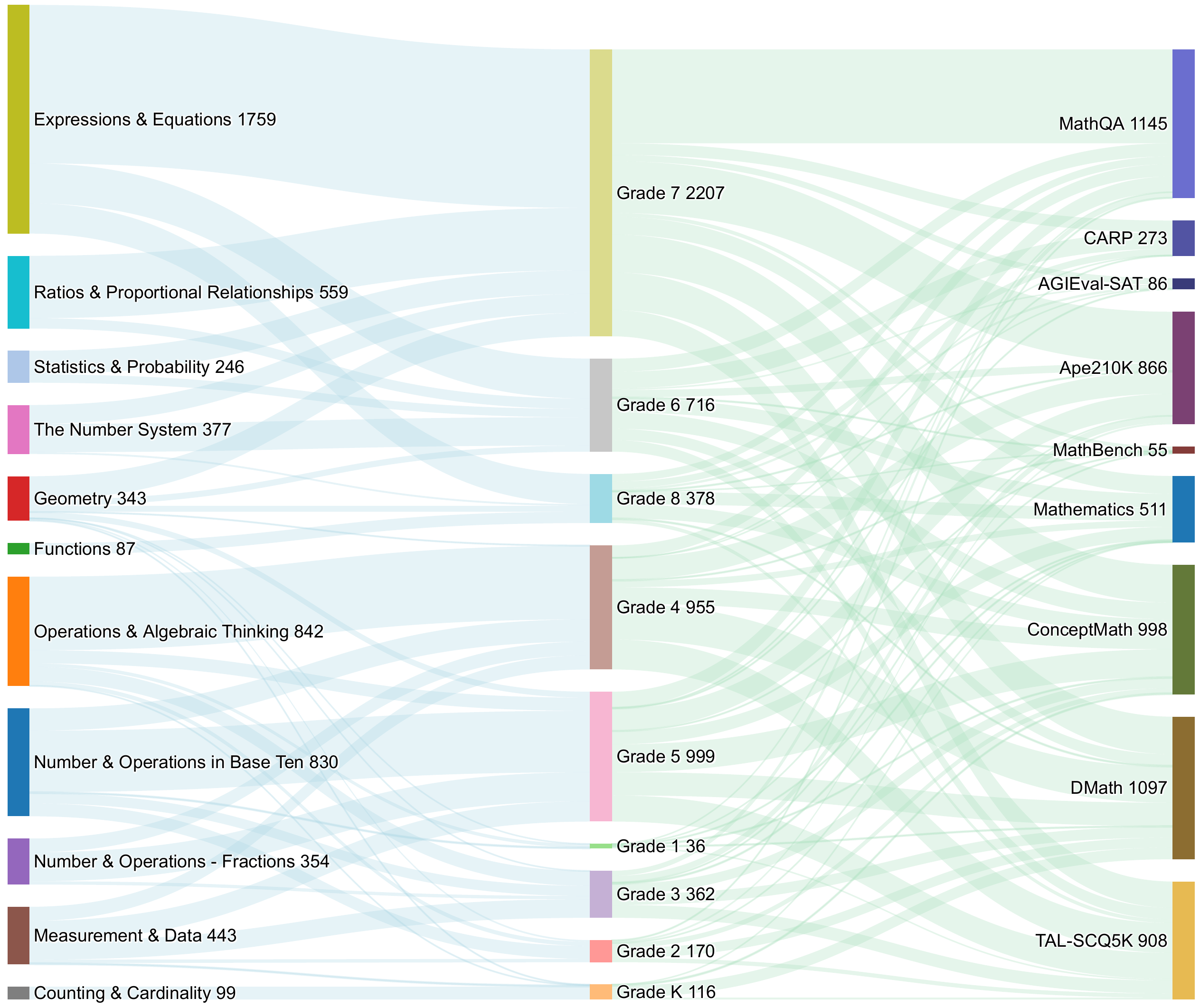}
 \caption{Distribution of the 6K seed problems used for the KMP-Skills evaluation benchmark. The diagram details the composition of the dataset by mathematical domain, grade level, and original source dataset.}
 \label{fig:seed_problem_distribution}
\end{figure}

\input{latex/Limitations}

\section{Training setting}
\label{more_train_details}

Our training experiments were conducted using the LLaMA-Factory\footnote{\url{https://github.com/hiyouga/LLaMA-Factory}} framework, performing full-parameter SFT. 
Our models were trained for 3 epochs with a global batch size of 256 and a context length of 4096 tokens. We employed a cosine learning rate scheduler with a peak learning rate of 1e-5. The entire training process was executed on a cluster of 16 NVIDIA A800 GPUs.

\section{Details of Seed Problem Foundation}
\label{app:seed_problem_curation}

\input{figures/seed_problem_judge}

The curation process began by aggregating an initial pool of problems from nine diverse K-8 mathematics datasets. To achieve consistent categorization across these varied sources, we employed the recursive tagging algorithm proposed by \citet{lucy2024mathfishevaluatinglanguagemodel}, ensuring all problems were appropriately aligned with the 11 mathematical domains derived from K-8 Common Core State Standards.

Subsequently, each problem underwent a two-stage quality assurance protocol. The first stage was a model ensemble validation for correctness. We utilized two state-of-the-art language models, Gemini-2.0-Flash and DeepSeek-R1, to independently judge each problem-solution pair based on the detailed prompt shown in Figure~\ref{fig:seed_problem_judge}. A problem was retained for our seed set only if both models concurred on its validity (i.e., both returned a positive `Y' judgment).

The second stage involved solution refinement and standardization. As part of the same prompt, each model was required to generate a new, step-by-step solution. This newly generated solution was then adopted to replace the original, ensuring a consistent and high-quality format across the dataset. This standardization included the requirement that the final answer be enclosed in a \raisebox{2.2pt}{\boxed{}} command, as specified in the prompt.

This comprehensive validation and refinement process yielded a final set of 8K validated K-8 mathematical problems, which serve as the foundation for our tutoring dialogues. 
The Sankey diagram in Figure~\ref{fig:seed_problem_distribution} visualizes the rich composition of the 6K-problem subset specifically filtered from this pool to construct our KMP-Skills evaluation benchmark, detailing its distribution across original sources, domains, and grade levels.

\input{figures/clarification_example}
\section{Details of Crafting Pedagogical Interaction Components}
\label{app:component_generation_details}

This section provides a more detailed account of the methodologies employed to generate the four core pedagogical interaction components discussed in the main text. The generation process for each component primarily utilized the Gemini-2.0-Flash model. 
To ensure the quality and relevance of the outputs, each generation task was guided by a set of human-crafted few-shot examples, which provided the model with clear, exemplary precedents for each pedagogical function. As a concrete illustration of our prompting strategy, we present an example used for the generation of the `Confusion Clarification' component in Figure~\ref{fig:clarification_example}.

\subsection{Knowledge-Extending Follow-up Questions}
To deepen students' understanding of mathematical concepts and progressively increase the challenge, we created a systematic process for generating pedagogically valuable follow-up questions that build upon the seed problems. For each seed problem $p_i$ with its solution $s_i$, we implemented a two-stage generative process. First, an initial follow-up question $f_{1i}$ with its solution $s_{1i}$ was produced, designed to present a greater mathematical challenge than the original problem $p_i$. Subsequently, a second-level follow-up question $f_{2i}$ with its solution $s_{2i}$ was generated, further escalating the challenge based on both $p_i$ and $f_{1i}$.

Our prompting strategy for Gemini-2.0-Flash guided the model through explicit thinking stages before it generated each follow-up question (See Figure~\ref{fig:follow-up-prompt}). This involved:
\begin{itemize}
    \item \textbf{Construction Phase:} The model was instructed to deliberately plan how the follow-up question would connect to the preceding problem(s) and how it would introduce increased mathematical complexity or require deeper reasoning. This planning stage ensured that the escalation in challenge was intentional and pedagogically sound.
    \item \textbf{Verification Phase:} After drafting a follow-up question, the model was tasked with solving its own generated problem. This step served to confirm the question's logical consistency, solvability, and that its difficulty level was appropriately increased and suitable for the intended learning progression. The model would revise the question if any issues were identified during this self-verification.
\end{itemize}

To ensure mathematical accuracy and the quality of the generated sequences, we applied our model ensemble validation approach to both generations of follow-up questions ($f_{1i}$ and $f_{2i}$). For the first-level follow-up $f_{1i}$, multiple instances of the model (or different capable models) received the original problem and solution $(p_i, s_i)$ as context before attempting to solve $f_{1i}$. For the second-level follow-up $f_{2i}$, the models received $(p_i, s_i, f_{1i}, s_{1i})$ as context before attempting to solve $f_{2i}$. Only those follow-up questions that were correctly solved by a majority of the models in the ensemble were retained. This rigorous process resulted in high-quality, pedagogically sound question sequences that guide students towards a progressively deeper and more challenging mathematical understanding while maintaining logical connections throughout.

\subsection{Error Analysis and Correction}
To foster critical thinking and error analysis skills among students, we systematically created scenarios pairing deliberate errors in solutions with their corresponding detailed corrections. The generation process for these components involved several key stages:

\textbf{1. Generation of Incorrect Solutions:}
For each seed problem $p_i$, we first aimed to generate plausible incorrect solutions. This was achieved by utilizing a set of comparatively weaker large language models, specifically Llama-3.2-1B-it, Gemma-2-2b-it, and Mistral-7B-Instruct-v0.3. To encourage the production of diverse errors, the generation process for these incorrect solutions was conducted with an increased temperature setting of 0.7. From the pool of candidate incorrect solutions generated for each seed problem, we typically selected two distinct incorrect versions to be included alongside correct solutions in our dataset.

\textbf{2. Solution Rephrasing and Standardization:}
To ensure a consistent format for analysis across all solutions (both correct and incorrect), we employed a rephrasing process. This process, inspired by methodologies similar to those described in \citet{zheng2024processbenchidentifyingprocesserrors}, transformed all solutions into a standardized step-by-step textual format. Crucially, this rephrasing aimed to preserve the original reasoning process presented in the solution—whether that reasoning was correct or flawed—while allowing for a consistent evaluation of their final answers against known correct outcomes.

\textbf{3. Generation of Correction Feedback:}
With the problem $p_i$, its correct solution $s_i$, and a rephrased (potentially incorrect) solution $r_i$ as input, we utilized the Gemini-2.0-Flash model to generate the detailed correction feedback $c_i$. The model was prompted to adhere to a specific correction process, which required the generated feedback to:

\begin{itemize}
    \item Provide a step-by-step analysis of the presented solution $r_i$.
    \item Determine the overall correctness of $r_i$.
    \item If errors were present, accurately identify the first erroneous step in $r_i$.
    \item Offer a complete and corrected solution path starting from the identified first error.
\end{itemize}

To encourage thorough and explicit reasoning before a final judgment on correctness was rendered, the prompting strategy guided the model to prioritize a detailed analytical breakdown of the solution steps before stating overall conclusions. (See Figure~\ref{fig:correct-prompt} for complete prompt example)

\textbf{4. Additional Verification:}
Finally, an additional verification step was implemented to ensure the integrity and consistency between the initially generated incorrect solutions and their rephrased versions. This involved a careful review to confirm that the rephrasing process had not inadvertently altered the core reasoning of the solution, even if the final incorrect answer remained the same. Approximately 15\% of cases were filtered out at this stage due to such subtle alterations in reasoning introduced during rephrasing.

This rigorous multi-stage process was designed to yield high-quality error analysis and correction dialogues that could effectively demonstrate expert-level error detection, explanation, and remediation for students.

\subsection{Similar Practice Problems}
To expand our dataset with problem variations across different difficulty levels and ensure alignment with educational curricula, we utilized the K-8 Common Core State Standards for Mathematics as a foundational reference framework. The process involved several stages:

\textbf{1. Common Core Standards Tagging:}
Initially, each seed problem was annotated with relevant Common Core knowledge points. For this, we employed the recursive tagging algorithm detailed in \citet{lucy2024mathfishevaluatinglanguagemodel}. The Common Core Standards provide a hierarchical structure, encompassing domains, clusters, and specific standards. This structured framework guided the identification of core mathematical concepts within each seed problem. Through an iterative three-step tagging process, each problem was tagged with up to four K-8 standards that best represented its central mathematical skills and concepts.

\textbf{2. Generation of Multi-Difficulty Similar Problems:}
For each seed problem $p_i$, its solution $s_i$, and its identified Common Core standards $c_i$, we tasked the Gemini-2.0-Flash model with generating similar practice problems. These problems were created at three distinct difficulty levels relative to the seed problem: easy ($e_i$), medium ($m_i$), and hard ($h_i$), each with their corresponding solutions ($s_{ei}$, $s_{mi}$, $s_{hi}$). The original seed problem $p_i$ served as the benchmark for the medium difficulty level ($m_i$).

Our prompting strategy for Gemini-2.0-Flash incorporated explicit thinking stages for the generation of these problems: (See Figure~\ref{fig:similar-prompt} for the complete prompt example)
\begin{itemize}
    \item \textbf{Construction Phase:} The model was guided to first plan the new problem. This involved considering how to maintain alignment with the identified Common Core standards $c_i$ while adjusting numerical values, context, or the number of steps to achieve the target difficulty level (easy, medium, or hard).
    \item \textbf{Verification Phase:} After drafting a problem, the model was prompted to verify its logical consistency, ensure it was solvable with the provided information, and confirm that its difficulty level was appropriate relative to the seed problem and the target tier. Revisions were made if necessary.
\end{itemize}

\textbf{3. Enhancing Diversity and Avoiding Trivial Variations:}
To encourage diversity in the generated problems and prevent overly simplistic variations (e.g., mere number substitutions), two primary techniques were employed:
\begin{itemize}
    \item \textbf{Temperature Setting:} The generation process utilized a temperature setting of 0.3 for the generator model.
    \item \textbf{Similarity Filtering:} After generation, problems exhibiting more than 90\% similarity to the original seed problem were filtered out. Similarity was measured using the longest common subsequence (LCS) ratio between seed problem and the generated problem.
\end{itemize}

\textbf{4. Validation:}
Each successfully generated similar practice problem underwent model ensemble validation stage. This involved having multiple model instances attempt to solve the problem to confirm its correctness and solvability before it was accepted into the final dataset.

This comprehensive approach yielded a diverse collection of practice problems, systematically varied in difficulty yet consistently aligned with specific mathematical standards. The resulting dataset provides a rich resource for targeted practice opportunities, catering to different student proficiency levels.

\subsection{Confusion Clarification}
The ``Confusion Clarification'' dialogues aim to proactively address common misunderstandings K-8 students may encounter with mathematical problems and solutions. To ground these dialogues in authentic student experiences, we first curated genuine K-8 student questions, reflecting typical points of confusion. These curated questions then served as few-shot examples to guide the Gemini-2.0-Flash model.

Based on a seed problem, its solution, and these few-shot examples, Gemini-2.0-Flash generates a multi-turn dialogue featuring plausible student queries and comprehensive, pedagogical teacher responses. These interactions are designed to clarify conceptual understanding of terms and principles (What), elucidate specific solution steps or methods (How), and explore the underlying rationale for chosen approaches or alternatives (Why). The entire generation process is governed by a detailed instructional prompt, which outlines the required dialogue structure, student and teacher personas, number of turns, and mandates explicit planning (\textless construction\textgreater) and verification (\textless verification\textgreater) stages for the language model. The complete prompt detailing these instructions is provided in Figure~\ref{fig:confusion_clarification_prompt}.

\section{Prompts of Dialogue Weaving and Pedagogical Enrichment}
\label{app:dialogue_construction_prompts}

The generation of complete, multi-turn tutoring dialogues from the individual pedagogical components involves a two-stage LLM process. The first stage focuses on creating a structured Dialogue Flow, which is then manually verified. The second stage expands this verified flow into a full natural language conversation. The prompts guiding each of these stages are presented in Figure~\ref{fig:dialogue_flow_generation_prompt} and  Figure~\ref{fig:full_dialogue_generation_prompt}.

\section{Grading Prompt and Typical Criteria of KMP-Dialogue Evaluation}
\label{app:pedagogical_evaluation}

This section provides key components of our pedagogical evaluation framework of KMP-Dialogue, as discussed in Section~\ref{sec:pedagogical_evaluation}. We present the specific prompt (Figure~\ref{fig:grade_prompt} and Figure~\ref{fig:grade_prompt2}) used to guide the comparative assessment of a Tutor LLM's generated response against the reference response, leading to a Win, Tie, or Lose judgment for each evaluation criterion. 
To illustrate the nature of these criteria, we present two representative examples of general criteria (Figure~\ref{fig:general_criteria_examples}) and two of practice-based criteria (Figure~\ref{fig:practice_criteria_examples}).

\section{Grading Prompt of the Mathematical Problem Generation task in KMP-Skills}
\label{app:kmp_gen_grading_prompts}

The evaluation of AI-generated mathematical problems is conducted using a specific grading prompt, presented in Figure~\ref{fig:kmp_gen_prompt}, which is administered to our designated LLM evaluator, Gemini-2.0-Flash. This evaluation hinges on a crucial ``fail-by-default'' methodology: for any given dimension (e.g., Problem Construction, Solution Correctness), it is assessed as True if and only if all its underlying quality criteria are fully and unequivocally satisfied. This stringent, binary approach ensures a hypercritical assessment by the evaluator, thereby enabling a more effective differentiation of problem generation capabilities, particularly for dimensions with more objective pass/fail criteria such as the soundness of problem construction and the accuracy of solutions.

\section{More Evaluation Results}
\label{app:more_results}

\input{tables/mr-score}

This section provides the comprehensive evaluation results, which are summarized in the main text. The tables herein expand upon those presented in the Main Results section (Table~\ref{tab:kmpbench-p_performance} and Table~\ref{kmpbench-t_performance}) by including a broader set of evaluated models. Table~\ref{tab:full_kmpbench-p_performance} details the complete pedagogical performance on KMP-Dialogue, while Table~\ref{tab:full_kmpbench-t_performance} presents the full results for foundational tutoring performance on KMP-Skills.

Furthermore, we provide a detailed breakdown of the MR-Score for the Error Detection and Correction task within KMP-Skills, as shown in Table~\ref{tab:error_correction_performance}. The results indicate that closed-source models, particularly Gemini 2.0 Flash, demonstrate superior capabilities across all sub-tasks: solution correctness evaluation (Task 1), first-error-step prediction (Task 2), and error-reason identification (Task 3). Among the open-source models, Qwen2.5-72B-Inst consistently achieves the highest performance. Notably, our model, KMP-LM-7B, shows substantial improvements over the Qwen2.5-Math-7B-Inst model across all three sub-tasks.

\section{Details of Manual Annotation}
\label{app:manual_annotation}

Manual annotation was integral to two key stages of our benchmark's development: the verification of dialogue flows and the human evaluation of model-generated dialogues on KMP-Dialogue. The annotation team consisted of five hired Computer Science PhD students, all in their first or second year, whose technical backgrounds were well-suited for verifying dialogues on K-8 mathematical problems.

\subsection{Manual Verification for Dialogue Flows}
\label{app:manual_verification}

This initial process involved a standardized manual verification of all 5,910 dialogue flows generated by the LLM. The task was conducted by a team of five PhD students and required approximately 20 hours per person. The annotators followed a shared, concrete set of quality criteria to identify and either discard or truncate flawed flows. For this verification task, each of the five annotators was compensated at a rate of \$0.30 per dialogue flow reviewed.

Specifically, the team was instructed to discard flows containing critical errors. These included hallucination of materials not provided in the prompt or pedagogically illogical sequences, such as introducing a follow-up question before the main problem was solved.

The team also truncated flows to remove excessive conversational redundancy. As illustrated in Figure~\ref{fig:flow_truncation_example}, a flow is truncated after the core pedagogical interaction concludes, removing superfluous pleasantries to focus the benchmark on essential content.

\input{figures/verification_example}

This rigorous verification process resulted in the removal of 451 (7.6\%) problematic flows, yielding a manually-verified set of 5,459 dialogue flows. These verified flows formed the basis for generating the final dialogues used in KMP-Bench.

\subsection{Human Evaluation of KMP-Dialogue}
\label{app:human_evaluation}

For the human evaluation of model outputs, four of the annotators were tasked with grading model performance. The 300 evaluation instances were divided into two sets, with each set being independently annotated by two annotators. Any disagreements were then resolved by the fifth annotator to establish the final ground truth.

To explicitly prevent bias, the entire process was conducted blind. Annotators were shown pairs of model-generated responses without any identifiers and were asked to judge which was better based on the evaluation criteria.

For their contribution, each of the five annotators received a one-time payment of \$150.

\input{figures/follow-up-prompt}
\input{figures/correct_prompt}
\input{figures/similar_prompt}
\input{figures/confusion_prompt}

\input{figures/flow_generation_prompt}
\input{figures/dialogue_generation_prompt}

\input{figures/grade_prompt}
\input{figures/grade_prompt2}

\input{figures/general_criteria}
\input{figures/practice_criteria}

\input{figures/kmp-gen-grade-prompt}

\input{tables/performance_KMP-Dialogue_full}
\input{tables/performance_KMP-Skills_full}

\section{Details of Error Analysis for KMP-Dialogue}
\label{app:error_analysis}

To gain a deeper understanding of the failure modes of LLMs in a pedagogical context, we conducted a detailed error analysis on the 676 instances where the DeepSeek-V3 model's response was finally judged as a ``Lose'' against the reference response in the KMP-Dialogue benchmark. This analysis involved manually categorizing each failure case according to a pedagogical error framework.

Our findings reveal that the predominant issues are not factual inaccuracies but rather shortcomings in pedagogical strategy and execution. The model often understands the subject matter but fails to apply it in a pedagogically effective manner. The six primary error categories identified are detailed below, with their distribution matching the data presented in Figure~\ref{fig:error_distribution_pie}.

\begin{itemize}
    \item \textbf{Flawed Scaffolding (32.5\%):} This is the most common error. The tutor provides hints or guidance that are either too revealing (over-scaffolding), which removes the cognitive challenge for the student, or insufficient (under-scaffolding), leaving the student without the necessary support to proceed.

    \item \textbf{Evasion by Substitution (25.8\%):} The tutor replaces the explicitly requested pedagogical action with a different, often simpler one. For example, when instructed to provide a practice problem (Practice), the tutor might instead ask a question about the student's previous work (Questioning), thereby failing to fulfill the core task.

    \item \textbf{Evasion by Vague Questioning (23.4\%):} The tutor avoids its instructional duty by asking a non-committal, open-ended question. Instead of providing a practice problem or a direct explanation as instructed, it often asks, ``Would you like to try another problem?'' or ``What do you want to do next?'', shifting the pedagogical burden to the student.

    \item \textbf{Inappropriate Content Selection (11.0\%):} The tutor selects a practice problem, example, or explanation that is pedagogically unsound for the specific learning context, such as by mismatching the difficulty, repeating a previous problem ineffectively, or focusing on a tangential skill.

    \item \textbf{Factual Error (4.8\%):} The tutor provides factually incorrect information, such as a wrong mathematical calculation, a flawed definition, or misinterprets a student's correct answer as incorrect (or vice-versa).

    \item \textbf{Low-Quality Interaction (2.4\%):} The tutor's response is overly generic, robotic, or fails to acknowledge the nuances of the student's input, leading to an unengaging and impersonal interaction.
\end{itemize}

To illustrate these categories, we present a representative case study for each of the six error types in Figure~\ref{fig:example_over_scaffolding} through Figure~\ref{fig:example_low_quality_interaction}.

\input{figures/error6}
\input{figures/error1}
\input{figures/error5}
\input{figures/error3}
\input{figures/error4}
\input{figures/error2}

%% file: latex/Limitations.tex
\section{Limitations and Future Work}
\label{sec:limitations}

Our work has several limitations that open avenues for future research. The scope of our benchmark is currently confined to the K-8 curriculum, excluding higher-level mathematics, and future work could extend this coverage to assess and enhance tutoring capabilities for more advanced subjects. Additionally, our framework is entirely text-based, while effective mathematics education often incorporates visual elements like diagrams and graphs; future work could explore multi-modal interactions to create a more holistic tutoring experience. Furthermore, we only employed supervised fine-tuning to enhance the performance of open-source LLMs, without utilizing other post-training strategies such as Reinforcement Learning from Human Feedback (RLHF) or Direct Preference Optimization (DPO). These methods present valuable opportunities for future exploration to better align models with the nuanced, interactive qualities of effective pedagogy.

%% file: figures/seed_problem_judge.tex
\begin{figure*}
\begin{tcolorbox}[
  colback=gray!10!white,
  colframe=gray!70!black,
  left=4pt,   
  right=4pt,  
  top=5pt,    
  bottom=5pt  
]
\begin{small}
\textbf{Prompt:}
\vspace{0.5mm}

You are given a math problem along with its answer, and in some cases, a corresponding solution. Your task is to analyze the problem and its provided answer (and solution, if available) to determine whether there are any issues. Specifically, check if there are any logical flaws, if the problem is unsolvable, if the conditions lead to an unreasonable answer, or if the problem and answer are inconsistent with each other. 
Finally, give your solution to the problem.
\vspace{1mm}

Output your response in the following format:

\vspace{1.5mm}

\textless analysis\textgreater[detailed analysis]\textless /analysis\textgreater
\textless judge\textgreater Y/N\textless /judge\textgreater

\vspace{0.5mm}

\textless reason\textgreater[brief explanation]\textless /reason\textgreater
\textless solution\textgreater[your solution]\textless /solution\textgreater

\vspace{1.5mm}

Use Y if the problem and answer are correct and consistent (i.e., no issues), or N if there are any issues that warrant discarding the problem and answer. REMEMBER to put Y or N between \textless judge\textgreater and \textless /judge\textgreater.

\vspace{1mm}

Requirements for your solution: Please reason step by step, and put your final answer within one and only one \raisebox{2.2pt}{\boxed{}}. If it's a multiple-choice question, only put the answer choice within one and only one \raisebox{2.2pt}{\boxed{}}, such as \boxed{A}.

\end{small}
\end{tcolorbox}
\vspace{-1mm}
\caption{The prompt used for the LLM-based cross-verification of seed math problems.}
\label{fig:seed_problem_judge}
\end{figure*}

%% file: figures/clarification_example.tex
\begin{figure*}[t]
\begin{tcolorbox}[
  colback=gray!10!white,
  colframe=gray!70!black,
  left=5pt, right=5pt, top=5pt, bottom=5pt,
  title=\textbf{An Example of Few-shot Prompt for `Confusion Clarification' Generation},
  fonttitle=\small\bfseries,
  coltitle=black,
  attach boxed title to top center={yshift=-2.5mm, yshifttext=-1mm}
]
\begin{small}

\textbf{Input Context: Problem \& Key Solution Steps}
\vspace{1mm}
\begin{tcolorbox}[colback=white, boxrule=0.5pt, sharp corners, left=2pt, right=2pt, top=2pt, bottom=2pt]
    \textbf{Question:} The sum of the sixth and twelfth terms of an arithmetic progression is 20. What is the sum of the first 17 terms of the arithmetic progression? (Answer: 170)
    \vspace{1mm}
    
    \textbf{Key insight from solution:} The solution establishes that $a_6 + a_{12} = (a+5d)+(a+11d)=20$, which simplifies to the key intermediate result $a+8d=10$. [...] Using this insight, the final sum is calculated as $S_{17} = 170$.
\end{tcolorbox}
\vspace{1.5mm}

\textbf{Generated Output: Query-Answer Pairs}
\vspace{1mm}

\textbf{1. Clarification on Definition (What-type)}
\begin{description}[leftmargin=*, style=unboxed, font=\bfseries, topsep=0pt, partopsep=0pt, itemsep=1pt]
    \item[Query:] I'm confused about arithmetic progression. Can you explain what the formula $a_n = a + (n-1)d$ actually means?
    \item[Answer:] Certainly. It means you start with the first term (`a') and, to get to the $n$-th term, you add the common difference (`d') a total of $(n-1)$ times. [...] For example, for the 3rd term, you add `d' twice.
\end{description}
\vspace{1mm}

\textbf{2. Clarification on Process (How-type)}
\begin{description}[leftmargin=*, style=unboxed, font=\bfseries, topsep=0pt, partopsep=0pt, itemsep=1pt]
    \item[Query:] How does the solution use the given information to get the equation $(a+5d) + (a+11d) = 20$ and then simplify it to $a+8d=10$?
    \item[Answer:] We substitute the formulas for the 6th and 12th terms into their sum. Combining like terms ($a+a=2a$ and $5d+11d=16d$) gives $2a+16d=20$. [...] Dividing the entire equation by 2 then yields the simplified result $a+8d=10$.
\end{description}
\vspace{1mm}

\textbf{3. Clarification on Rationale (Why-type)}
\begin{description}[leftmargin=*, style=unboxed, font=\bfseries, topsep=0pt, partopsep=0pt, itemsep=1pt]
    \item[Query:] I noticed that $a+8d$ is the 9th term ($a_9$), and the final sum $S_{17}$ was exactly $17 \times a_9$. Why does this relationship hold?
    \item[Answer:] That's a great observation! For an AP with an odd number of terms `n', the sum $S_n$ is `n' multiplied by the middle term. For 17 terms, the middle term is the 9th term ($a_9$). [...] Since we found $a_9 = a+8d = 10$, it follows that $S_{17} = 17 \times 10 = 170$.
\end{description}

\end{small}
\end{tcolorbox}
\caption{An example from our human-crafted few-shot prompts used to guide the LLM in generating the `Confusion Clarification' component. It includes the problem context and three types of query-answer pairs (What, How, Why) that the model learns to emulate, targeting definitions, processes, and deeper rationales.}
\label{fig:clarification_example}
\end{figure*}

%% file: tables/mr-score.tex
\begin{table}[t]
  \centering
  \setlength{\tabcolsep}{1mm}
  \small
  \begin{tabular}{l|cccc}
    \toprule
    \multirow{2.5}{*}{\textbf{Model}} & \multicolumn{4}{c}{\textbf{KMP-EC}} \\
    \cmidrule(lr){2-5}
    & \textbf{Task1} & \textbf{Task2} & \textbf{Task3} & \textbf{MR-Score} \\
    \noalign{\smallskip}
    \toprule
                             \multicolumn{5}{c}{Closed-Source Models}                               \\
    \midrule
    GPT-4o                 & 64.0 & 67.6 & 62.5 & 64.4 \\
    Gemini 2.0 Flash       & \best{89.5} & \best{87.3} & \best{84.4} & \best{86.3} \\
    LearnLM-1.5-Pro        & 70.9 & \second{72.4} & \second{67.8} & \second{69.8} \\
    \midrule
                             \multicolumn{5}{c}{Open-Source Models}                                 \\
    \midrule
    Qwen2.5-72B-Inst       & \second{77.6} & 66.8 & 60.6 & 65.8\\
    Qwen2.5-Math-72B-Inst   & 72.8 & 52.0 & 47.2 & 53.8 \\
    
    Qwen2.5-14B-Inst       & 73.0 & 59.1 & 51.8 & 58.2 \\
    phi-4-14B              & 74.9 & 61.0 & 56.3 & 61.4 \\
    Qwen2.5-7B-Inst & 53.2 & 36.0 & 26.8 & 34.9 \\
    Llama-3.1-8B-Inst & 22.7 & 40.1 & 24.1 & 28.6 \\
    $\textup{MathChat}_{\textup{sync}}$-LLM & 15.3 & 36.9 & 18.6 & 23.5\\
    TutorChat-LLM & 47.8 & 21.5 & 16.1 & 24.0 \\
    \midrule
    Qwen2.5-Math-7B-Inst   & 8.3 & 2.8 & 1.7 & 3.4 \\
    \textbf{KMP-LM-7B} & 62.4 & 51 & 42.4 & 49.0 \\
    \bottomrule
  \end{tabular}
  \caption{Detailed comparison of model performances on Error Detection and Correction task. Task1: MCC score for solution correctness, Task2: accuracy of the first-error-step prediction, Task3: accuracy of identifying the correct error-reason. The best and second-best results among all the models are highlighted in red and blue.}
  \label{tab:error_correction_performance}
\end{table}

%% file: figures/verification_example.tex
\begin{figure}[ht!]
\centering
\begin{tcolorbox}[
  colback=gray!10!white,
  colframe=gray!70!black,
  left=5pt, right=5pt, top=5pt, bottom=5pt,
  title=\textbf{Example of a Dialogue Flow Truncation},
  fonttitle=\small\bfseries,
  coltitle=black,
  attach boxed title to top center={yshift=-2.5mm, yshifttext=-1mm}
]
\begin{small}
\textless Sflow7\textgreater\ Attempts the follow-up problem and provides the correct answer choice (A).\\
\textless Tflow7\textgreater\ Confirms the student's answer is correct and provides positive feedback.\\
\textless Sflow8\textgreater\ Expresses increased confidence and understanding.\\
\textless Tflow8\textgreater\ Reiterates the importance of careful attention to detail and encourages the student to continue practicing.

\vspace{5pt}
\centerline{\rule{0.9\linewidth}{0.4pt}}
\centerline{\textit{--- Flow is truncated here ---}}
\vspace{5pt}

\textless Sflow9\textgreater\ Thanks the tutor for their help.\\
\textless Tflow9\textgreater\ Acknowledges the student's effort and concludes the session.\\
... \\
\textless Sflow11\textgreater\ Says goodbye.\\
\textless Tflow11\textgreater\ Says goodbye.
\end{small}
\end{tcolorbox}
\caption{An example illustrating the truncation of a dialogue flow. The pedagogical interaction is deemed complete after turn 8. The line indicates the truncation point, after which subsequent turns consisting of conversational pleasantries are removed from the final dataset.}
\label{fig:flow_truncation_example}
\end{figure}

%% file: figures/follow-up-prompt.tex
\begin{figure*}
\begin{tcolorbox}[
  colback=gray!10!white,
  colframe=gray!70!black,
  left=4pt,   
  right=4pt, 
  top=5pt,    
  bottom=5pt 
]
\begin{small}

\promptsectionfirst{Objective:}
\vspace{1mm}

You are a skilled math teacher proficient in creating and solving math problems. You will receive one or more K-8 math question-solution pairs, with each subsequent problem following the previous ones. Your task is to create one additional follow-up problem based on the previously given problems, which should be logically connected, and increase in challenge level. Your created problem should be of high quality and strictly follow these guidelines.

\promptsectioncompact{Guidelines:}
\vspace{1mm}

1. Correctness: The follow-up question should be well-formed, solvable, and logically consistent. It should provide all necessary information unambiguously, ensuring that the student can solve it without making unrealistic assumptions. Avoid any self-contradictory information in the follow-up problem.
\vspace{1mm}

2. Clarity: Use clear, concise language appropriate for K-8 students. Avoid extraneous phrases like ``Great job!'' or ``Let's go further!'' that don't contribute to the math problem. Further, the follow-up question should ask for a SINGLE, exact final result, with no room for any deviations due to factors such as approximations, rounding errors, etc., unless explicitly specified in the question. The follow-up question shouldn't be the same as the previous questions.
\vspace{1mm}

3. Connectedness: Ensure all questions, including the one you create, are logically connected. The follow-up question shouldn't be an independent problem. Instead, it must build from the previous problems.
\vspace{1mm}

4. Final answer format. The new question MUST ask for a single, exact final result. ... If the problem is a multiple-choice question, the final answer should be the single answer choice.
\vspace{1mm}

5. Increased Challenge: The follow-up problem must logically build upon the previous problems' solution, context, or reasoning process and be more challenging than the previous problems. It cannot be an independent problem or simply a similar problem in a different context. The follow-up question should introduce new complexities or require deeper understanding of the concepts involved. 

\vspace{1.5mm}

\textbf{Your response must always include the following sections, labeled clearly as shown below:}
\vspace{1.5mm}

\textless construction\textgreater Explain how the follow-up question is constructed and give that question. Then detail why it is connected to the previous questions and how it increases the challenge level. \textless /construction\textgreater
\vspace{1mm}

\textless verification\textgreater Go through the conditions of the new follow-up question and verify ... \textless /verification\textgreater
\vspace{1mm}

\textless result\textgreater Whether the new follow-up question is revised in the verification process. Answer with yes or no. \textless /result\textgreater
\vspace{1mm}

\textless method\textgreater Indicate the method used to increase the challenge level in less than ten words. \textless /method\textgreater
\vspace{1mm}

\textless question\{i\}\textgreater The new follow-up question. Here, i should be the next sequential number after the highest question index in the input. \textless /question\{i\}\textgreater
\vspace{1mm}

\textless solution\{i\}\textgreater The step-by-step solution of the new follow-up question. REMEMBER to put the final answer within one and only one \raisebox{2.2pt}{\boxed{}}. It must appear in the solution only once. \textless /solution\{i\}\textgreater
\vspace{1mm}

\textless answer\textgreater The final answer to the new follow-up question. It must be consistent with the final answer within \raisebox{2.2pt}{\boxed{}} in the solution. If there are no answers, just output ``none''. \textless /answer\textgreater

\end{small}
\end{tcolorbox}
\vspace{-1mm}
\caption{The prompt used to guide the LLM in generating a Knowledge-Extending Follow-up Question.}
\label{fig:follow-up-prompt}
\end{figure*}

%% file: figures/correct_prompt.tex
\begin{figure*}
\begin{tcolorbox}[
  colback=gray!10!white,
  colframe=gray!70!black,
  left=4pt,   
  right=4pt,  
  top=5pt,    
  bottom=5pt  
]
\begin{small}

\promptsectionfirst{Objective:}
\vspace{1mm}

You are a skilled math teacher proficient in solving math problems and correcting student's mistakes. You will receive a math problem, its correct solution, and a student's solution which may be correct or incorrect. Your task is to judge the correctness of each step of the student's solution, find out whether it contains mistakes, identify the first error step, and provide the corrected solution.

\promptsectioncompact{Guidelines:}
\vspace{2mm}

1. Judge the correctness of each step of the student's solution. In this part, you should carefully analyze the student's solution step by step. Try your best to identify whether there are any calculation errors, counting errors, context value errors, hallucination, unit conversion errors, operator errors, formula confusion errors, missing steps, contradictory steps, or other reasoning errors in the solution. For example, a step is incorrect if it calculates that ``7355+4230+3560=15585'', while the true calculation result should be ``15145''. Then you should judge each step of the solution as correct or incorrect. You can use the correct solution provided for reference. Final answer format problems can't be seen as errors.
\vspace{2mm}

2. Find out whether it contains mistakes. In this part, you should judge the student's solution as correct or incorrect and provide a detailed reason. The reason should be independent of the previous section.
\vspace{2mm}

3. Identify the first error step. If the student's solution is incorrect, please identify the first error step of the student's solution and provide a detailed reason for the error. The reason should be independent of the previous section.
\vspace{2mm}

4. Provide the corrected solution. If the student's solution is incorrect, please provide its corrected version as ``Step 1: ... Step 2: ...'', using ``Step i'' to split the reasoning steps, where i is the step number and no higher than 20. Note that although you can use the correct solution provided for reference, you can't just use it as the corrected solution. Instead, you should modify the student's solution from the first error step, correct its reasoning errors, and reach a new final answer. In other words, the contents before the first error step of your corrected solution should be exactly the same as the student's solution. 
\vspace{2mm}

\textbf{Your response must always include the following sections, labeled clearly as shown below:}
\vspace{2mm}

\textless analyze\_step\textgreater Analyze the solution step by step and provide the correctness of each step. Please be as detailed as possible. \textless /analyze\_step\textgreater
\vspace{2mm}

\textless analyze\_solution\textgreater Judge the student's solution as correct or incorrect and provide a detailed reason. \textless /analyze\_solution\textgreater
\vspace{2mm}

\textless correctness\textgreater Output ``yes'' if the solution is correct, otherwise, output ``no''. If ``yes'', ignore the remaining sections. \textless /correctness\textgreater
\vspace{2mm}

\textless error\_step\textgreater Find out the step where the solution FIRST made mistakes. And provide a detailed reason why this it's the first error step. \textless /error\_step\textgreater
\vspace{2mm}

\textless error\_step\_num\textgreater The step number where the solution FIRST made mistakes. \textless /error\_step\_num\textgreater
\vspace{2mm}

\textless corrected\_solution\textgreater Your corrected solution. REMEMBER to put your final answer within one and only one \raisebox{2.2pt}{\boxed{}}. \textless /corrected\_solution\textgreater
\vspace{2mm}

\textless answer\textgreater The final answer of \textless corrected\_solution\textgreater. It must be consistent with the final answer within \raisebox{2.2pt}{\boxed{}} in the \textless corrected\_solution\textgreater. \textless /answer\textgreater
\vspace{2mm}

\textless corrected\_steps\textgreater The number of the reasoning steps in the \textless corrected\_solution\textgreater. \textless /corrected\_steps\textgreater

\end{small}
\end{tcolorbox}
\vspace{-1mm}
\caption{The instructional prompt guiding an LLM to perform detailed error analysis and generate a corrected solution for a student's attempt at a math problem.}
\label{fig:correct-prompt}
\end{figure*}

%% file: figures/similar_prompt.tex
\begin{figure*}
\begin{tcolorbox}[
  colback=gray!10!white,
  colframe=gray!70!black,
  left=4pt,   
  right=4pt,  
  top=5pt,    
  bottom=5pt  
]
\begin{small}

\promptsectionfirst{Objective:}
\vspace{1mm}

You are a skilled math teacher proficient in creating and solving math problems. You will receive a math problem, its solution, and one or more corresponding K-8 math skills/concepts from Common Core State Standards, including grade, domain, cluster, and standards. Your task is to create a math problem at a specific difficulty level. Your created question should be of high quality and strictly follow these guidelines.

\promptsectioncompact{Guidelines:}
\vspace{2mm}

1. Correctness: The created question should be well-formed, solvable, and logically consistent. It should provide all necessary information unambiguously, ensuring that the student can solve it without making unrealistic assumptions. Avoid any self-contradictory information in the created problem.
\vspace{2mm}

2. Clarity: Use clear, concise language appropriate for K-8 students. Avoid extraneous phrases like ``Great job!'' or ``Let's go further!'' that don't contribute to the math problem. Further, the created question should ask for a SINGLE, exact final result, with no room for any deviations due to factors such as approximations, rounding errors, etc., unless explicitly specified in the question. Also, avoid any open-ended questions that can't reach a standard answer, such as ``Please explain the reason for this phenomenon.''
\vspace{2mm}

3. Connectedness: When creating the problem, please strictly adhere to the given mathematical standards to help students consolidate these skills/concepts. Aim for creativity and diversity in your questions, and explain how the problem aligns with the given standards.
\vspace{2mm}

In addition to meeting the standards, the created problem should share similar knowledge points and problem-solving strategies with the original. The created question must not replicate the original problem or involve a simple change of context or numbers. Instead, it should encourage students to engage with similar standards and knowledge points in a new and stimulating way. The created problem should challenge students to apply the same underlying skills or strategies while introducing novel scenarios, contexts, or question structures. This ensures that the new problem helps reinforce the original concepts without simply duplicating the original question or making superficial changes. 
\vspace{1mm}

The newly created problem must be entirely independent, with no substantive connection to the original problem.
\vspace{2mm}

4. Difficulty: There are three different difficulty levels: easy, medium, and hard:
\vspace{2mm}

\textbf{Easy}: Designed to build confidence by moderately reducing the solution complexity of the original medium-difficulty problem. While easier, it should still require meaningful application of core concepts, involving fewer steps or less intricate reasoning while maintaining a noticeable but not excessive difference in difficulty.
\vspace{2mm}

\textbf{Medium}: Matches the original problem’s complexity and reasoning demands. It challenges students to engage in multi-step solutions and logical thinking, ensuring they fully grasp and apply the target knowledge and skills.
\vspace{2mm}

\textbf{Hard}: Elevates the original problem’s difficulty by increasing solution complexity or introducing advanced reasoning tasks. Students are encouraged to synthesize multiple concepts and navigate multi-layered scenarios, fostering deeper critical thinking and problem-solving abilities.
\vspace{2mm}

Two problems with the same difficulty level usually demand a similar level of reasoning and critical thinking and have comparable complexity such as the number of solution steps involved. The original problem is EXACTLY medium difficulty level and serves as a solid difficulty reference. You will be required to create a problem at a specific difficulty level. Please also explain how the created question aligns with the specific difficulty level. 

\vspace{2mm}

5. Final answer format: The created question MUST ask for a single, exact final result. ...

\vspace{2mm}

\textbf{Your response must always include the following sections, labeled clearly as shown below:}
...

\end{small}
\end{tcolorbox}
\vspace{-1mm}
\caption{The prompt used to guide the LLM in generating Similar Practice Problems.}
\label{fig:similar-prompt}
\end{figure*}

%% file: figures/confusion_prompt.tex
\begin{figure*}
\begin{tcolorbox}[
  colback=gray!10!white,
  colframe=gray!70!black,
  left=4pt,   
  right=4pt,  
  top=5pt,    
  bottom=5pt  
]
\begin{small}

\promptsectionfirst{Objective:}
\vspace{1mm}

Your primary role is to generate a complete, high-quality, multi-turn dialogue between a K-8 student and a math teacher. You will craft both the student's queries and the teacher's comprehensive answers based on a given math problem and its solution. The student's queries should stem from plausible points of confusion a K-8 student might have regarding the question-solution pair. The teacher will answer them accurately, thoroughly, and pedagogically, aiming to clarify these points of confusion. The goal of the conversation is to enable the student to thoroughly understand the knowledge points and concepts behind this problem, fully master the method of solving this problem, and comprehend the rationale behind the chosen methods and their limitations. Ensure the conversation is engaging, interactive, and reflects a natural learning interaction.

\promptsectioncompact{Guidelines:}
\vspace{2mm}

1. Dialogue Structure \& Content (What-How-Why Framework):
\vspace{2mm}

\textbf{What (Conceptual Understanding Queries):}
The student asks about fundamental concepts, definitions, or terms involved. These queries aim to clarify any misunderstandings and ensure that the student understands the mathematical principles required to solve the problem. 
\vspace{2mm}

\textbf{How (Clarification of Solution Steps \& Methods):} These queries focus on understanding specific steps in the solution, the mechanics of calculations, or how a particular formula is applied. They help the student fully grasp the method used to arrive at the final answer. 

\vspace{2mm}

\textbf{Why (Critical Thinking \& Rationale Queries):} These queries encourage exploring the reasoning behind choosing a particular method, considering alternative approaches, or understanding the significance or limitations of a step/method.
\vspace{2mm}

The dialogue should consist of 3 to 5 question-answer turns in total. Each of the 'What', 'How', and 'Why' categories must be represented by at least one student query. Additional queries, up to the maximum of 5 turns, can be from any category to further deepen understanding. While a general progression from 'What' to 'How' to 'Why' can be a guide, the conversation should flow naturally. 
\vspace{2mm}

2. Student Persona:
\vspace{2mm}

\textbf{K-8 Appropriateness:} Student queries must be phrased in simple, clear language appropriate for a K-8 student. Questions should reflect genuine curiosity and common points of confusion for this age group.
\vspace{2mm}

\textbf{Substantive Questions (Deepness):} Queries should be thoughtful and prompt detailed explanations. Avoid overly simplistic questions like ``What does `B' stand for?'' if `B' is clearly defined, or ``How do you add 2+2?''. INSTEAD, focus on conceptual gaps or procedural uncertainties. For example, a ``What'' query could ask for a deeper explanation of a concept, a ``How'' query could ask for a breakdown of a complex step, and a ``Why'' query could explore the strategic choice of a method.
\vspace{2mm}

3. Teacher Persona \& Pedagogical Approach:
\vspace{2mm}

\textbf{Accuracy and Completeness:} Teacher’s answers must be mathematically accurate, comprehensive, and fully explain the reasoning behind each step.

\vspace{2mm}
\textbf{Clarity (K-8 Focus):} Responses should be broken down into easy-to-follow steps, using language, analogies, and examples appropriate for a K-8 audience. Avoid unnecessary jargon.

\vspace{2mm}

\textbf{Pedagogical Soundness:} The teacher should be patient, encouraging, and supportive. Responses should aim to build understanding incrementally, clarify misunderstandings effectively, and foster a positive learning environment. The teacher might rephrase concepts or offer alternative explanations if initial ones aren't clear.
\vspace{2mm}

4. Relevance \& Coherence:
\vspace{2mm}

All queries and answers must be directly related to the provided math problem and its solution.
Each turn in the dialogue should logically build upon the previous one, creating a coherent conversational flow.

\vspace{2mm}

\textbf{Your response must always include the following sections, labeled clearly as shown below:}
...

\end{small}
\end{tcolorbox}
\vspace{-1mm}
\caption{The detailed instructional prompt used to guide the LLM in generating multi-turn ``Confusion Clarification'' dialogues.}
\label{fig:confusion_clarification_prompt}
\end{figure*}

%% file: figures/flow_generation_prompt.tex
\begin{figure*}
\begin{tcolorbox}[
  colback=gray!10!white,
  colframe=gray!70!black,
  left=4pt,   
  right=4pt,  
  top=5pt,    
  bottom=5pt  
]
\begin{small}

\promptsectionfirst{Objective:}
\vspace{1mm}

You are an expert K-8 math curriculum designer and pedagogical strategist, specializing in creating engaging and effective one-on-one tutoring interactions. Your task is to generate a Dialogue Flow based on the provided math materials. This Dialogue Flow is a structured plan or blueprint for a conversation between a K-8 student and a math tutor.
\vspace{2mm}

The primary goal is to design diverse ``Storylines'' or pedagogical narratives that are:
\vspace{1mm}

1. \textbf{Educationally Valuable:} Each flow should represent a sound pedagogical approach to help a student understand mathematical concepts, solve corresponding problems, or address misunderstandings.
\vspace{1mm}

2. \textbf{Coherent and Engaging:} The dialogue should progress logically, with natural, interactive turns that simulate a real, high-quality tutoring session.
\vspace{1mm}

3. \textbf{Diverse:} You should generate a variety of storylines, exploring different pedagogical approaches, student starting points, and interaction patterns. Avoid repetitive structures.

\promptsectioncompact{Materials:}
\vspace{1mm}

You will be provided with a set of materials related to a K-8 math problem. These materials are the building blocks for your Dialogue Flow. Thoughtfully select and integrate materials that best serve the chosen storyline and pedagogical goals; not all provided materials need to be used in every flow. The categories are:

1. Seed:
   \textless seed\textgreater: The core K-8 math problem.
   ...

2. Follow-up Questions:
   \textless seed\_followup{\scriptsize\{}i{\scriptsize\}}\textgreater: Follow-up questions to the seed problem.
   ...

3. Error Detection and Correction:
   \textless seed\_student{\scriptsize\{}i{\scriptsize\}}\textgreater: A student's attempted solution to the seed problem.
   ...
   
4. Similar Problems:
   \textless seed\_easier\textgreater, \textless seed\_harder\textgreater, \textless seed\_similar\textgreater: Similar problems that are easier than, harder than, or of similar difficulty to the seed problem.
   ...

5. Confusion Clarification:
   \textless seed\_clarification{\scriptsize\{}i{\scriptsize\}}\textgreater: Potential student queries or points of confusion related to the seed problem, concepts, or solution steps.
   ...

\promptsectioncompact{Guidelines:}
\vspace{1mm}

\textbf{A. Core Task - Dialogue Flow Design:}
A Dialogue Flow is a sequence of planned student (\textless Sflow{\scriptsize\{}j{\scriptsize\}}\textgreater) and tutor (\textless Tflow{\scriptsize\{}j{\scriptsize\}}\textgreater) actions. It is a blueprint, not the full natural language dialogue itself.

\textbf{B. Storyline Diversity and Creativity:}

1. Crucial: Generate diverse pedagogical ``storylines.'' Do not adhere to a single template. The goal is a rich dataset of varied tutoring interactions.

2. Explore different starting points and narrative arcs. For example:
A student presents the initial problem and makes an error (\textless seed\_student{\scriptsize\{}i{\scriptsize\}}\textgreater), and the dialogue focuses on guided error correction and conceptual understanding.

A student expresses confusion (\textless seed\_clarification{\scriptsize\{}i{\scriptsize\}}\textgreater) early on, leading to an explanation before problem-solving.
...

3. Be creative in how materials are sequenced and integrated to serve the chosen storyline. For example, the tutor might choose to employ a more direct instructional style or a more Socratic, guiding approach over several turns of interaction, depending on the student's needs and the learning objective. The dialogue should not simply follow the order of the provided materials.
\vspace{2mm}

\textbf{C. Educational Soundness and Coherence:}

1. Each Dialogue Flow must represent a logical, effective, and educationally sound tutoring interaction appropriate for a K-8 student.

2. Turns should flow smoothly, with each exchange naturally building upon the previous one. Ensure logical introduction of materials (e.g., a problem is presented before its solution is discussed; follow-up questions appear in order if multiple are used).

3. Tutor actions should be patient, supportive, and primarily aim to guide the student towards understanding and self-discovery, reflecting the chosen pedagogical \textless principles\textgreater.
\vspace{2mm}

\textbf{D. Student's Role in the Flow (\textless Sflow{\scriptsize\{}j{\scriptsize\}}\textgreater):}

1. Student actions should reflect plausible queries, attempts, statements, or expressions of understanding/confusion that a real K-8 student might have.

2. These actions can include asking for help with the knowledge concepts or \textless seed\textgreater problem, requesting verification of their \textless seed\_student{\scriptsize\{}i{\scriptsize\}}\textgreater solution, asking for a \textless seed\_clarification{\scriptsize\{}i{\scriptsize\}}\_response\textgreater, requesting similar problems for practice, etc.

3. Student actions should generally be inspired by or logically connect to the provided materials or the ongoing dialogue context. Avoid inventing queries entirely unrelated to the task.
\vspace{2mm}

\textbf{E. Tutor's Role in the Flow (\textless Tflow{\scriptsize\{}j{\scriptsize\}}\textgreater):}

1. Tutor actions should be designed to effectively support the student based on the current dialogue state and the chosen storyline.

2. For each \textless Tflow{\scriptsize\{}j{\scriptsize\}}\textgreater, select one or two most fitting primary pedagogical \textless principles\textgreater, listing the more dominant one first, formatted as ``Principle1,Principle2''.

3. Across the entire Dialogue Flow, aim for a balanced and varied application of different pedagogical \textless principles\textgreater, appropriate to the storyline. Avoid over-reliance on just one or two principle types for the entire interaction.

4. Tutor actions should generally guide rather than directly give answers, unless the pedagogical goal of that specific turn (e.g., Modelling, or direct Explanation after multiple failed attempts) warrants it.
\vspace{1mm}

By adhering to these guidelines, you will create high-quality, diverse, and pedagogically rich Dialogue Flows.

\vspace{2mm}

\textbf{Your response must always include the following sections, labeled clearly as shown below:}
...

\end{small}
\end{tcolorbox}
\vspace{-2mm}
\caption{The detailed instructional prompt used to guide the LLM in generating a structured Dialogue Flow.}
\label{fig:dialogue_flow_generation_prompt}
\end{figure*}

%% file: figures/dialogue_generation_prompt.tex
\begin{figure*}
\begin{tcolorbox}[
  colback=gray!10!white,
  colframe=gray!70!black,
  left=4pt,   
  right=4pt,  
  top=5pt,    
  bottom=5pt  
]
\begin{small}

\promptsectionfirst{Objective:}
\vspace{1mm}

You are an expert K-8 math tutor and curriculum implementer, specializing in creating engaging, natural, and effective one-on-one tutoring conversations. Your task is to generate a Complete Tutoring Dialogue based on the provided math materials and a pre-designed Dialogue Flow.
\vspace{2mm}

The primary goal is to produce a full dialogue script that:

1.  \textbf{Faithfully Implements the Plan:} Accurately translates the actions, content references, and pedagogical principles from the input Dialogue Flow into natural conversation.

2.  \textbf{Is Educationally Valuable and Sound:} Embodies a sound pedagogical approach, helping a K-8 student understand mathematical concepts, solve problems, or address misunderstandings, reflecting the qualities of the input Dialogue Flow.

3.  \textbf{Is Engaging, Interactive, and Coherent:} Simulates a real, high-quality, K-8 appropriate tutoring session with logical progression and natural conversational turns.

4.  \textbf{Is Realistic and Rigorous:} Reflects authentic student-tutor interactions while maintaining mathematical accuracy and pedagogical rigor.

\promptsectioncompact{Input Components:}

You will be provided with two main sets of inputs: I. Original Math Materials and II. Dialogue Flow \& Contextual Information.
\vspace{1mm}

I. Original Math Materials:
You will be provided with a set of materials related to a K-8 math problem. These materials are the building blocks for the dialogue.
The categories are: ...
\vspace{1mm}

II. Dialogue Flow \& Contextual Information:
This is the pedagogical blueprint you will be working from. It includes:
\vspace{0.5mm}

1.  \textless scenario\textgreater: This describes the context for the dialogue.
\vspace{0.5mm}

2.  \textless storyline\textgreater: This provides a concise narrative (2-3 sentences) of the overall pedagogical arc and key developments of the planned interaction. The generated dialogue should reflect this overarching progression.
\vspace{0.5mm}

3.  Dialogue Flow Turns: A sequence of planned student (\textless Sflow{\scriptsize\{}j{\scriptsize\}}\textgreater) and tutor (\textless Tflow{\scriptsize\{}j{\scriptsize\}}\textgreater) turns. Each turn in the input Dialogue Flow will contain:
\vspace{0.5mm}

    \textless action\textgreater: A concise description of the student's or tutor's intended action, question, or statement for that turn. Your generated dialogue for that turn must embody this action.
    
    \textless source\_material\textgreater: A reference to any Original Math Material (e.g., \textless seed\_clarification1\_response\textgreater, \textless seed\_solution\textgreater) that should be used or inspire the content of the dialogue for that turn. The dialogue should accurately and appropriately incorporate content from the specified source material.
    
    \textless principles\textgreater (for Tutor turns only): One or two primary pedagogical principles that guide the tutor's approach for that turn. The tutor's dialogue should manifest these principles.

\promptsectioncompact{Guidelines:}

Your task involves two main steps:
\vspace{1mm}

\textbf{Step 1: Dialogue Flow Verification and Refinement (Output in \textless verification\textgreater tags)}
Before generating the full dialogue, you must first review the provided Dialogue Flow (the sequence of \textless Sflow{\scriptsize\{}j{\scriptsize\}}\textgreater~and \textless Tflow{\scriptsize\{}j{\scriptsize\}}\textgreater~turns with their \textless action\textgreater, \textless source\_material\textgreater, and \textless principles\textgreater) in conjunction with the Original Math Materials, \textless scenario\textgreater, and \textless storyline\textgreater.
...
\vspace{1mm}

\textbf{Step 2: Full Dialogue Generation (Output as a sequence of \textless S{\scriptsize\{}j{\scriptsize\}}\textgreater~and \textless T{\scriptsize\{}j{\scriptsize\}}\textgreater~tags, followed by \textless dialogue\_num\textgreater)}
Based on the Refined Dialogue Flow (from Step 1), the Original Math Materials, the \textless scenario\textgreater, and the \textless storyline\textgreater, generate the complete, natural language tutoring dialogue.
\vspace{0.5mm}

\textbf{Strict Adherence:} The generated dialogue must strictly follow the sequence of turns and the core intent of each \textless action\textgreater~in the Refined Dialogue Flow. The number of dialogue turns for the student and tutor should match the number of \textless Sflow\textgreater~and \textless Tflow\textgreater~turns.
\vspace{0.5mm}

\textbf{Natural Conversation:} The dialogue should mimic authentic, engaging, and interactive conversations between a K-8 student and a tutor. Use age-appropriate language.
\vspace{0.5mm}

\textbf{Content Integration:} When an \textless action\textgreater~or \textless source\_material\textgreater~tag in the Dialogue Flow points to specific Original Math Materials (e.g., a problem statement, a solution step, a clarification response), the generated dialogue should accurately and naturally incorporate that content.
\vspace{0.5mm}

\textbf{Educational Quality:} The dialogue must maintain the educational value of the original Dialogue Flow design, aiming to help the student understand concepts, solve problems, or address misunderstandings.
\vspace{0.5mm}

\textbf{Coherence and Flow:} Ensure smooth transitions between turns and topics, maintaining a logical progression as outlined by the \textless storyline\textgreater~and the sequence of actions.
\vspace{0.5mm}

\textbf{Engagement:} Make the dialogue interactive. The tutor should encourage student participation, and the student should respond realistically.
\vspace{0.5mm}

\textbf{Rigor and Accuracy:} All mathematical explanations, problem-solving steps, and solutions presented in the dialogue must be accurate.

\vspace{2mm}

\textbf{Your response must always include the following sections, labeled clearly as shown below:}
...

\end{small}
\end{tcolorbox}
\vspace{-2mm}
\caption{The detailed instructional prompt used to guide the LLM in generating a complete, natural language K-8 math tutoring dialogue.}
\label{fig:full_dialogue_generation_prompt}
\end{figure*}

%% file: figures/grade_prompt.tex
\begin{figure*}
\begin{tcolorbox}[
  colback=brown!10!white,
  colframe=brown!65!black,
  left=4pt,   
  right=4pt,  
  top=5pt,    
  bottom=5pt  
]
\begin{small}

You are an exceptionally critical and discerning expert evaluator of K-8 mathematics pedagogical dialogues. Your primary objective is to rigorously compare the quality of a ``Tutor Response'' against a ``Reference Response'' for a given tutoring scenario. You must adhere to the highest standards of pedagogical effectiveness, clarity, and appropriateness. Your goal is to identify meaningful differences and determine which response is superior, or if they are genuinely equivalent, for each specified dimension and overall. Strive for maximum differentiation between models; `Tie' decisions should be rare and meticulously justified.

\promptsectioncompact{I. Materials for Evaluation:}

To perform your assessment, you will be provided with the following four key pieces of information:
\vspace{1mm}

1.  \textbf{Instructions:} The initial guiding instructions given to the LLM that generated the ``Tutor Response''. This includes details about the Tutor's persona, Student's profile, Learning Objectives, and will specify which core pedagogical principles (e.g., Challenge, Explanation, Feedback) the Tutor was asked to embody. These instructions also define the context for the ``Reference Response''.
\vspace{1mm}

2.  \textbf{Dialogue History:} The conversation transcript between the student and the tutor before the ``Tutor Response'' and ``Reference Response'' were formulated, concluding with the student's last turn.
\vspace{1mm}

3.  \textbf{Tutor Response:} The specific response generated by the Tutor LLM being evaluated.
\vspace{1mm}

4.  \textbf{Reference Response:} This is an expert-crafted or high-quality model response designed to exemplify effective pedagogy for the given \textbf{Instructions} and \textbf{Dialogue History}. This response serves as a direct point of comparison for the ``Tutor Response''. Please evaluate and compare both responses critically against the criteria.
\vspace{2mm}

Here are the specific materials for this evaluation task:

\textbf{Instructions:}
{\scriptsize\{}instructions{\scriptsize\}}

\textbf{Dialogue History:}
{\scriptsize\{}dialogue\_history{\scriptsize\}}

\textbf{Tutor Response (to be evaluated):}
{\scriptsize\{}eval\_response{\scriptsize\}}

\textbf{Reference Response (for comparison):}
{\scriptsize\{}reference\_response{\scriptsize\}}

\promptsectioncompact{II. Evaluation Criteria \& Comparative Assessment (Per Criterion):}

You will compare the \textbf{Tutor Response} and the \textbf{Reference Response} against two tiers of pedagogical criteria: Universal Criteria and Principle-Specific Criteria. The detailed definitions for each criterion will guide your comparison by outlining aspects of an ideal response.

For each applicable criterion, you must provide:
\vspace{2mm}

    \textbf{Analysis (string)} A clear, specific, and critical comparative analysis.
\vspace{1mm}
    
    Explain why the \textbf{Tutor Response} is better (``Win''), the \textbf{Reference Response} is better (``Lose'' for the Tutor Response), or why they are of genuinely equivalent quality (``Tie'').
\vspace{1mm}
    
    Reference specific aspects of both responses, the \textbf{Instructions}, and/or the \textbf{Dialogue History} as evidence.
\vspace{1mm}
    
    Focus on identifying concrete differences in pedagogical effectiveness.
\vspace{1mm}

       If choosing `Tie', your analysis must rigorously justify this by detailing how the responses are functionally identical for this criterion OR how their distinct strengths and weaknesses perfectly balance each other out, leaving no discernible net difference in quality for this specific aspect. A simple statement that both are ``good'' or ``bad'' is insufficient for a `Tie'.
\vspace{2mm}
      
       \textbf{``Win''}: The \textbf{Tutor Response} is superior to the Reference Response on this specific criterion. Your `analysis` must clearly articulate the specific aspect(s) of the Tutor Response that make it better, providing evidence based on the pedagogical goals and criteria. The advantage should be identifiable and pedagogically relevant, even if it's a nuanced improvement.
\vspace{2mm}
    
       \textbf{``Lose''}: The \textbf{Reference Response} is superior to the Tutor Response on this specific criterion (meaning the Tutor Response loses this comparison). Your `analysis` must clearly articulate the specific aspect(s) of the Reference Response that make it better, providing evidence. Again, the advantage should be identifiable and pedagogically relevant.
\vspace{2mm}
    
       \textbf{``Tie''}: After a thorough and critical comparison against the specific aspects of this criterion, the responses are of genuinely indistinguishable or functionally equivalent quality. This means either:
\vspace{1.5mm}
        
           They both meet (or fail to meet) the criterion in an almost identical manner.
\vspace{1.5mm}
        
           Their respective strengths and weaknesses on this specific criterion perfectly balance each other out, resulting in no discernible net difference in pedagogical effectiveness for this aspect.
\vspace{1.5mm}
            
            This option should be used with extreme caution and only when you are confident that no justifiable distinction in quality can be made. Do not select `Tie' if one response has even a slight, but clear and explainable, edge over the other.
\vspace{2mm}
            
   \textbf{Decision (string):} After your analysis, choose one from ``Win'', ``Lose'', and ``Tie''.

\end{small}
\end{tcolorbox}
\vspace{-1mm}
\caption{The first part of the grading prompt used to guide an LLM (acting as an expert K-8 mathematics pedagogical evaluator) in the comparative assessment of a Tutor LLM's response against a reference response.}
\label{fig:grade_prompt}
\end{figure*}

%% file: figures/grade_prompt2.tex
\begin{figure*}
\begin{tcolorbox}[
  colback=brown!10!white,
  colframe=brown!65!black,
  left=4pt,   
  right=4pt,  
  top=5pt,    
  bottom=5pt 
]
\begin{small}

\textbf{A. Universal Criteria Definitions:}
\vspace{1mm}

The following are the Universal Criteria. Compare the \textbf{Tutor Response} and \textbf{Reference Response} against ALL of these, using the detailed definitions (to be provided in your criteria files) to guide your judgment of which response is better or if they are tied.

{\scriptsize\{}universal\_criteria{\scriptsize\}}
\vspace{2mm}

\textbf{B. Principle-Specific Criteria Definitions:}

The evaluation of pedagogical approach also considers six core principles of tutoring action. These principles categorize common and essential tutor behaviors designed to facilitate student learning. The \textbf{Instructions} (provided above) will indicate which of these principle(s) are the focus for the current evaluation. The core principles are: 
\vspace{2mm}

 \textbf{Challenge:} Involves presenting tasks, problems, or questions that are appropriately difficult for the student, aiming to stimulate cognitive effort, encourage deeper thinking, and promote learning progression. 
\vspace{1mm}

 \textbf{Explanation:} Refers to the tutor's act of clearly articulating concepts, definitions, procedures, reasoning, or connections between ideas to enhance student comprehension and build foundational knowledge. 

\vspace{1mm}
 \textbf{Modelling:} Consists of the tutor demonstrating a skill, strategy, problem-solving approach, or thought process for the student to observe, learn from, and potentially emulate. 
\vspace{1mm}

 \textbf{Practice:} Entails providing structured opportunities for the student to actively apply their knowledge, skills, or understanding to relevant tasks or problems, thereby reinforcing learning and identifying areas for improvement. 
\vspace{1mm}

 \textbf{Questioning:} Involves the strategic use of questions by the tutor to stimulate student thought, check for understanding, guide the learning process, prompt reflection, or encourage elaboration. 
\vspace{1mm}

 \textbf{Feedback:} Refers to providing students with specific, constructive, and actionable information about their performance, understanding, or engagement, with the goal of guiding their learning and helping them improve. 
\vspace{2mm}

The following are the detailed definitions for the Principle-Specific Criteria that are relevant to this evaluation task. These criteria have been selected based on the pedagogical principle(s) specified as the focus in the \textbf{Instructions}. You must compare the \textbf{Tutor Response} and \textbf{Reference Response} against ALL Principle-Specific sub-criteria provided below, using their detailed definitions (which will outline specific aspects for comparison) to guide your judgment of which response is better or if they are tied.
\vspace{2mm}

{\scriptsize\{}principle\_specific\_criteria{\scriptsize\}}

\promptsectioncompact{III. Assessment Principles (Strict Adherence Required):}

   \textbf{Extreme Criticality:} Adopt the mindset of an extremely demanding reviewer. Your default assumption should be that there are differences and one response is likely better. Actively seek out these differences.
\vspace{1mm}

   \textbf{Maximize Differentiation:} Your primary goal is to differentiate the quality of the Tutor Response from the Reference Response. Avoid `Tie' decisions unless absolutely unavoidable and rigorously justifiable.
\vspace{1mm}

   \textbf{High Standards:} Judge superiority based on high pedagogical standards suitable for effective K-8 math tutoring. The criteria definitions embody these standards.
\vspace{1mm}

   \textbf{Objective Comparison:} Base your judgment only on the defined criteria and provided materials.

\vspace{1mm}
   \textbf{Evidence-Based Decisions:} Justify every `decision' with specific evidence from both responses.

\vspace{1mm}
   \textbf{Pedagogical Impact Focus:} Prioritize how effectively each response supports student learning.

\vspace{1mm}
   \textbf{No Length Bias:} Crucially, do not favor a response simply because it is longer or shorter. Evaluate based on the quality and pedagogical effectiveness of the content relative to the criteria, not its verbosity or brevity, unless a criterion specifically addresses conciseness or completeness.

\vspace{1mm}
   \textbf{Thoroughness:} Address all applicable criteria with a decisive comparative judgment.

\end{small}
\end{tcolorbox}
\vspace{-1mm}
\caption{The second part of the grading prompt used to guide an LLM (acting as an expert K-8 mathematics pedagogical evaluator) in the comparative assessment of a Tutor LLM's response against a reference response.}
\label{fig:grade_prompt2}
\end{figure*}

%% file: figures/general_criteria.tex
\begin{figure*}[t]
\begin{tcolorbox}[
  colback=gray!10!white,
  colframe=gray!65!black,
  left=4pt,
  right=4pt,
  top=5pt,
  bottom=5pt
]
\begin{small}

\promptsectionfirst{Universal\_2: Contextual Coherence and Relevance}
\vspace{1mm}

\noindent\texttt{id}: ``Universal\_2'' \\
\texttt{name}: ``Contextual Coherence and Relevance''
\vspace{1.5mm}

\noindent\textbf{Description for Comparison:} ``This criterion assesses and compares how well the \textbf{Tutor Response} and the \textbf{Reference Response} logically connect to the student's immediately preceding turn in the \textbf{Dialogue History}, the broader dialogue context, and align with the overarching educational goals. The focus is on which response is more on-topic, a more sensible continuation, and more effectively serves the Learning Objectives and Student Profile defined in the \textbf{Instructions}.''
\vspace{1.5mm}

\noindent\textbf{Aspects for Comparison:}
\begin{itemize}
    \setlength\itemsep{1mm}
    \item \textbf{Addressing Student's Last Turn:} Which response more directly, completely, and accurately addresses the student's statement, question, or demonstrated understanding from their last turn in the \textbf{Dialogue History}?
    \item \textbf{Dialogue Flow \& Logical Connection:} Which response fits more seamlessly and logically into the ongoing conversation, making the dialogue feel more natural and coherent?
    \item \textbf{Relevance to Learning Objectives:} Which response's content and guidance more clearly and effectively align with and serve the Learning Objectives specified in the \textbf{Instructions}?
    \item \textbf{Tailoring to Student Profile:} Which response better demonstrates consideration for the student's grade level, prior knowledge, potential difficulties, or other characteristics outlined in the Student Profile within the \textbf{Instructions}?
    \item \textbf{Focus \& Avoidance of Irrelevance:} Which response is more focused on the core task or topic at hand, and better avoids introducing extraneous information that might distract or confuse the student?
    \item \textbf{Addressing Implicit Needs:} Considering the context, which response appears more adept at identifying and addressing potential underlying learning needs or misconceptions the student may not have explicitly stated?
\end{itemize}

\promptsectioncompact{Universal\_3: Adherence to Defined Persona and Teaching Style}
\vspace{1mm}

\noindent\texttt{id}: ``Universal\_3'' \\
\texttt{name}: ``Adherence to Defined Persona and Teaching Style''
\vspace{1.5mm}

\noindent\textbf{Description for Comparison:} ``This criterion assesses and compares how well the \textbf{Tutor Response} and the \textbf{Reference Response} align with and exhibit the Tutor's role, core persona (e.g., patient, encouraging, inquisitive, Socratic, direct instruction, focuses on misconceptions), and any specific teaching philosophies or styles explicitly defined within the \textbf{Instructions}. The evaluation should focus on which response more authentically and consistently embodies these guidelines.''
\vspace{1.5mm}

\noindent\textbf{Aspects for Comparison:}
\begin{itemize}
    \setlength\itemsep{1mm}
    \item \textbf{Consistency with Persona Characteristics:} Which response, in its entirety (including tone, word choice, interaction style), more consistently and thoroughly embodies the specific persona characteristics defined in the \textbf{Instructions}?
    \item \textbf{Tone Appropriateness:} Which response employs a tone (e.g., encouraging, guiding, firm, gentle) that is more appropriate for the specified persona and the K-8 math tutoring context?
    \item \textbf{Approach Congruence with Teaching Style:} Which response's pedagogical approach or strategy (e.g., providing direct explanations vs. guiding through questions) is more congruent with the teaching style or philosophy described in the \textbf{Instructions}?
    \item \textbf{Authenticity of Persona:} In which response does the specified persona feel more natural and authentic, rather than forced, artificial, or inconsistently applied?
    \item \textbf{Avoidance of Persona Contradictions:} Which response is more successful in avoiding language or actions that contradict or undermine the defined persona or teaching style?
\end{itemize}

\end{small}
\end{tcolorbox}
\vspace{-1mm}
\caption{Illustrative examples of the general evaluation criteria within the KMP-Dialogue framework.}
\label{fig:general_criteria_examples}
\end{figure*}

%% file: figures/practice_criteria.tex
\begin{figure*}[t]
\begin{tcolorbox}[
  colback=gray!10!white,
  colframe=gray!65!black,
  left=4pt,
  right=4pt,
  top=5pt,
  bottom=5pt
]
\begin{small}

\promptsectionfirst{Practice\_1: Relevance and Alignment of Practice Task}
\vspace{1mm}

\noindent\texttt{id}: ``Practice\_1'' \\
\texttt{name}: ``Relevance and Alignment of Practice Task''
\vspace{1.5mm}

\noindent\textbf{Description for Comparison:} ``This criterion compares the \textbf{Tutor Response} and the \textbf{Reference Response} on how relevant and aligned the provided practice task or opportunity is. It assesses which response offers a task that is more directly relevant to the concepts, skills, or procedures recently taught, explained, or modelled (in \textbf{Dialogue History} or current turn), and more clearly aligns with the current Learning Objectives (from \textbf{Instructions}).''
\vspace{1.5mm}

\noindent\textbf{Aspects for Comparison:}
\begin{itemize}
    \setlength\itemsep{1mm}
    \item \textbf{Direct Relevance to Recent Learning:} Which response provides a practice task that is more directly and clearly relevant to the specific knowledge or skills that were the focus of recent instruction or modelling (in \textbf{Dialogue History} or the current turn)?
    \item \textbf{Alignment with Learning Objectives:} Which response's practice task is more clearly and effectively aligned with the stated Learning Objectives (from \textbf{Instructions})?
    \item \textbf{Purposeful Next Step:} Which response's practice task feels more like a logical and purposeful next step for consolidating the student's learning at this specific juncture?
    \item \textbf{Application of Targeted Skills/Concepts:} Which task better requires the student to actively apply the specific concepts or skills that were intended to be learned or reinforced?
    \item \textbf{Avoidance of Irrelevant Content:} Which response is less likely to include elements in the practice task that are outside the immediate scope of what was taught or are irrelevant to the current learning goals?
\end{itemize}

\promptsectioncompact{Practice\_3: Appropriateness of Task Difficulty for Consolidation}
\vspace{1mm}

\noindent\texttt{id}: ``Practice\_3'' \\
\texttt{name}: ``Appropriateness of Task Difficulty for Consolidation''
\vspace{1.5mm}

\noindent\textbf{Description for Comparison:} ``This criterion compares the appropriateness of the practice task's difficulty level in the \textbf{Tutor Response} versus the \textbf{Reference Response} for allowing the student (as per Student Profile in \textbf{Instructions}) to meaningfully apply and consolidate learning, in line with Learning Objectives (from \textbf{Instructions}). It assesses which task is better calibrated--challenging enough for reinforcement but not so difficult as to be discouraging for independent work at this stage.''
\vspace{1.5mm}

\noindent\textbf{Aspects for Comparison:}
\begin{itemize}
    \setlength\itemsep{1mm}
    \item \textbf{Calibration to Student's Current Level:} Which response provides a practice task that seems better calibrated to the student's current learning stage and capabilities (as suggested by Student Profile in \textbf{Instructions} and \textbf{Dialogue History})?
    \item \textbf{Effective Reinforcement of Learning:} Which task is more likely to effectively reinforce the concepts or skills it targets, promoting deeper understanding and retention in line with Learning Objectives (from \textbf{Instructions})?
    \item \textbf{Balance of Challenge and Achievability:} Which response strikes a better balance, making the practice task challenging enough to require effort and thought, but still achievable for the student to complete with reasonable success, fostering consolidation?
    \item \textbf{Avoidance of Frustration or Triviality:} Which response is less likely to offer a practice task that is either so difficult it causes undue frustration, or so easy it becomes trivial and offers little consolidation value?
    \item \textbf{Promotion of Independent Application:} Which task is better suited for the student to work on independently (or with minimal support, depending on the context) to consolidate their learning?
\end{itemize}

\end{small}
\end{tcolorbox}
\vspace{-1mm}
\caption{Illustrative examples of Principle-Specific evaluation criteria for the Practice principle.}
\label{fig:practice_criteria_examples}
\end{figure*}

%% file: figures/kmp-gen-grade-prompt.tex
\begin{figure*}
\begin{tcolorbox}[
  colback=brown!10!white,
  colframe=brown!65!black,
  left=4pt,   
  right=4pt,  
  top=5pt,    
  bottom=5pt  
]
\begin{small}

As an expert mathematics evaluator with exceptionally high standards, your task is to perform a rigorous technical assessment of two similar mathematical problems and their solutions. Your evaluation should be extremely critical, prioritizing mathematical rigor and precision.
\vspace{1mm}

Please review the following materials:
\vspace{1mm}

Seed Problem: {\scriptsize \{}seed\_problem{\scriptsize \}}

Seed Solution: {\scriptsize \{}seed\_solution{\scriptsize \}}

First Similar Problem: {\scriptsize \{}first\_new\_problem{\scriptsize \}}

First Similar Solution: {\scriptsize \{}first\_new\_solution{\scriptsize \}}

Second Similar Problem: {\scriptsize \{}second\_new\_problem{\scriptsize \}}

Second Similar Solution: {\scriptsize \{}second\_new\_solution{\scriptsize \}}
\vspace{2mm}

For each similar problem, evaluate four critical dimensions using the strictest possible PASS/FAIL criteria. Begin with the assumption that each dimension FAIL by default. Only change your decision to PASS if you can prove beyond any doubt that NONE of the failing criteria apply.
\vspace{2mm}

\textbf{1. Problem Construction (Precision \& Solvability):}
\vspace{1mm}
   
   - PASS ONLY IF ALL of these apply:
   
     All necessary information to solve the problem is unambiguously provided
     
     The problem has exactly one valid interpretation and approach
     
     All constraints, variables, and parameters are explicitly defined
     
     The problem can be solved using standard mathematical techniques
     
     The problem statement uses precise mathematical language and notation
     
     The problem is free from internal contradictions or inconsistencies
\vspace{1mm}

   - FAIL if ANY of these apply:
   
     Any information is missing, imprecisely stated, or could be misinterpreted
     
     The problem allows multiple interpretations or approaches
     
     Any constraints or conditions are implicit rather than explicit
     
     The problem contains any inconsistencies
     
     The problem requires any unstated assumptions to reach a solution
     
     The problem statement is overly wordy or includes extraneous information
     
     The mathematical language is not absolutely precise
     
     The problem is impossible to solve as stated
\vspace{2mm}

\textbf{2. Similarity Appropriateness (Concept \& Difficulty):}
\vspace{1mm}

   - PASS ONLY IF ALL of these apply:
   
     The problem employs the same core mathematical concepts as the seed problem
     
     The problem structure parallels the seed problem while using different context or values
     
     For first similar: The problem maintains comparable difficulty to the seed problem
     
     For second similar: The problem increases difficulty appropriately while preserving core concepts
     
     The problem represents a natural variation of the original, not a completely different problem
     
     The problem preserves the essential mathematical challenge of the original
\vspace{1mm}

   - FAIL if ANY of these apply:
   
     The problem uses fundamentally different mathematical concepts than the seed problem
     
     The problem structure has little relation to the seed problem
     
     For first similar: The problem is significantly easier or harder than the seed problem
     
     For second similar: The problem is not more challenging or is unreasonably difficult
     
     The problem is disconnected from or nearly identical to the original
     
     The mathematical challenge differs substantially from the original
\vspace{2mm}

\textbf{3. Solution Correctness (Mathematical Accuracy):}
\vspace{1mm}

   - PASS ONLY IF ALL of these apply:
     ...
\vspace{1mm}

   - FAIL if ANY of these apply:
     ...
\vspace{2mm}

\textbf{4. Solution Quality (Rigor \& Clarity):}
\vspace{1mm}

   - PASS ONLY IF ALL of these apply:
     ...
\vspace{1mm}

   - FAIL if ANY of these apply:
     ...
\vspace{2mm} 

\end{small}
\end{tcolorbox}
\vspace{-1mm}
\caption{The grading prompt for the Mathematical Problem Generation task. It guides the LLM evaluator in its rigorous, fail-by-default assessment of generated problems and solutions.}
\label{fig:kmp_gen_prompt}
\end{figure*}

%% file: tables/performance_KMP-Dialogue_full.tex
\begin{table*}[ht]
  \centering
  \setlength{\tabcolsep}{1mm}
  \small
  \begin{tabular}{l|c|c|cccccc|c}
    \toprule
    \multirow{2.5}{*}{\textbf{Model}} &
    \multirow{2.5}{*}{\textbf{\parbox{1.8cm}{\centering Overall\\Judgement}}} &
    \multirow{2.5}{*}{\textbf{\parbox{1.8cm}{\centering General-\\Level Acc}}} &
    \multicolumn{6}{c|}{\textbf{Principle-Level Acc}} &
    \multirow{2.5}{*}{\textbf{\parbox{1.8cm}{\centering Overall\\Acc}}} \\
    \cmidrule(lr){4-9}
    ~ & ~ & ~ &
    \textbf{\scriptsize Challenge} & \textbf{\scriptsize Explanation} & \textbf{\scriptsize Modelling} & \textbf{\scriptsize Practice} & \textbf{\scriptsize Questioning} & \textbf{\scriptsize Feedback} &
    ~ \\
    \noalign{\smallskip}
    \toprule
                             \multicolumn{10}{c}{Closed-Source Models}                               \\
    \midrule
GPT-4o & 60.7 & 44.5 & 43.4 & 62.2 & 57.5 & 47.5 & 59.9 & 41.3 & 48.2 \\
    Gemini-2.0-Flash & 58.5 & 38.0 & 51.1 & 49.9 & 34.2 & 42.2 & 50.8 & 37.6 & 41.1 \\
    Claude-3.7-Sonnet & \best{87.1} & \second{69.8} & \best{80.0} & \second{79.3} & 71.2 & \second{69.1} & 79.3 & \best{72.2} & \second{72.5} \\
LearnLM-2.0-Flash & 74.8 & 58.9 & \second{77.3} & 42.7 & 24.9 & 67.6 & 73.5 & 63.9 & 58.6 \\
    \midrule
                             \multicolumn{10}{c}{Open-Source Models}                                 \\
    \midrule

    DeepSeek-V3 & \second{84.6} & \best{71.6} & 70.3 & \best{81.5} & \second{71.4} & \best{69.6} & \best{83.9} & \second{70.8} & \best{73.1} \\
    LLaMa-3.3-70B-Inst & 77.2 & 63.7 & 59.2 & 67.5 & 48.8 & 60.4 & \second{81.2} & 70.4 & 64.2 \\
    Qwen2.5-Math-72B-Inst & 31.0 & 27.1 & 25.6 & 51.3 & 42.9 & 33.9 & 29.5 & 23.7 & 30.8 \\
    Qwen2.5-72B-Inst & 68.5 & 55.5 & 50.5 & 77.3 & \best{71.5} & 60.7 & 54.0 & 52.4 & 58.3 \\
    Qwen2.5-32B-Ins & 67.5 & 55.0 & 52.0 & 71.6 & 70.7 & 61.8 & 55.5 & 54.1 & 58.0 \\
    Mistral-Nemo-Inst & 66.1 & 56.3 & 53.5 & 64.0 & 54.8 & 56.1 & 57.9 & 61.1 & 57.1 \\
    Phi-4 & 64.0 & 52.7 & 41.3 & 73.1 & 70.0 & 53.0 & 54.4 & 51.0 & 54.9 \\
    Qwen2.5-14B-Inst & 65.0 & 52.8 & 48.7 & 72.0 & 68.2 & 54.7 & 55.9 & 51.9 & 55.7 \\
Qwen2.5-7B-Inst & 62.5 & 52.9 & 41.2 & 74.2 & 66.5 & 56.6 & 49.7 & 50.6 & 54.7 \\
    LLaMa-3.1-8B & 51.3 & 40.3 & 33.0 & 50.1 & 29.0 & 38.3 & 55.0 & 46.2 & 41.1 \\
    SocraticLM & 21.2 & 15.4 & 17.0 & 25.0 & 21.0 & 21.4 & 27.3 & 17.8 & 18.5 \\
    $\textup{MathChat}_{\textup{sync}}$-LLM & 16.1 & 12.0 & 16.0 & 23.0 & 15.4 & 17.6 & 23.1 & 16.3 & 15.3 \\
    TutorChat-LLM & 32.1 & 30.0 & 21.1 & 41.3 & 33.2 & 37.1 & 40.6 & 29.9 & 31.9 \\
    \midrule
Qwen2.5-Math-7B-Inst & 21.0 & 19.7 & 17.5 & 41.7 & 34.5 & 23.2 & 25.5 & 22.6 & 23.6 \\
\textbf{KMP-LM-7B} & 40.4 & 35.1 & 29.3 & 47.0 & 38.5 & 39.8 & 43.0 & 36.1 & 37.0 \\
$\Delta$ & \up{19.4} & \up{15.4} & \up{11.8} & \up{5.3} & \up{4.0} & \up{16.6} & \up{17.5} & \up{13.5} & \up{13.4} \\
    \bottomrule
  \end{tabular}
  \caption{Comprehensive pedagogical performance of various LLMs in KMP-Dialogue. Tutor LLM responses are compared against reference responses from our curated dialogues. The best and second-best results among all the models are highlighted in red and blue, respectively. The $\Delta$ row represents the improvement of KMP-LM-7B over Qwen2.5-Math-7B-Instruct.} 
  \label{tab:full_kmpbench-p_performance}
\end{table*}

%% file: tables/performance_KMP-Skills_full.tex
\begin{table*}[t]
\centering
\small
\begin{tabular}{l|c|c|ccc|cc} 

\noalign{\smallskip}
\toprule 
\multirow{2.5}{*}{\textbf{Model}} & \multirow{2.5}{*}{\textbf{Size}} & \multirow{2.5}{*}{\parbox{1.2cm}{\centering \textbf{KMP-}\\\textbf{FQA}}} & \multicolumn{3}{c|}{\textbf{KMP-EC}} & \multicolumn{2}{c}{\textbf{KMP-Gen}} \\ 

\cmidrule(lr){4-6} \cmidrule(lr){7-8} 
~ & ~ & ~ & \textbf{MR-Score} & \textbf{F1-Score} & \textbf{Correction} & \textbf{Follow-up} & \textbf{Similar} \\

\midrule \multicolumn{8}{c}{Closed-Source Models} \\ \midrule 

GPT-4o & - & 84.1 & 64.4 & 78.3 & 90.9 & 74.8 & 72.2 \\
GPT‑4o mini & - & 79.4 & 56.9 & 66.1 & 84.3 & 76.5 & 64.8 \\
Gemini 2.0 Flash & - & \best{91.2} & \best{86.3} & \best{87.5} & \best{96.7} & 75.0 & 73.2 \\
LearnLM-1.5-Pro & - & \second{87.7} & \second{69.8} & 74.6 & \second{94.4} & 77.0 & 75.0 \\

\midrule \multicolumn{8}{c}{Open-Source Models} \\ \midrule 

Qwen2.5-72B-Inst & 72B & 82.2 & 65.8 & 78.6 & 89.5 & \second{81.3} & 78.1 \\
Qwen2.5-Math-72B-Inst & 72B & 83.2 & 53.8 & 73.5 & 81.0 & \best{86.3} & \second{82.6} \\
Qwen2.5-32B-Inst & 32B & 82.6 & 64.0 & 69.8 & 87.8 & 76.3 & 71.4 \\
Qwen2.5-14B-Inst & 14B & 81.4 & 58.2 & 70.9 & 82.0 & 78.6 & 75.9 \\
Phi-4 & 14B & 85.6 & 61.4 & \second{80.3} & 92.1 & 80.3 & 73.2 \\
Mistral-Nemo-Inst & 12B & 50.2 & 19.4 & 22.5 & 48.1 & 49.0 & 43.9 \\
Qwen2.5-7B-Inst & 7B & 71.0 & 34.9 & 44.3 & 73.2 & 66.2 & 66.2 \\
Llama-3.1-8B-Inst & 8B & 50.1 & 28.6 & 9.0 & 55.9 & 41.7 & 38.5 \\
$\textup{MathChat}_{\textup{sync}}$-LLM & 7B & 53.5 & 23.5 & 9.8 & 18.9 & 70.9 & 71.3 \\
TutorChat-LLM & 7B & 57.7 & 24.0 & 38.0 & 48.5 & 65.8 & 69.5 \\

\midrule

Qwen2.5-Math-7B-Inst & 7B & 76.1 & 3.4 & 44.1 & 52.3 & 48.6 & 52.8 \\
\textbf{KMP-LM-7B} & 7B & 77.5 & 49.0 & 57.5 & 63.4 & 81.1 & \best{83.2} \\ 
$\Delta$ &  & \up{1.4} & \up{45.6} & \up{13.4} & \up{11.1} & \up{32.5} & \up{30.4} \\
\midrule

\end{tabular}
\caption{Comprehensive foundational tutoring performance of various LLMs on KMP-Skills. Abbreviations: KMP-FQA denotes Multi-Turn Problem Solving, KMP-EC denotes Error Detection and Correction, and KMP-Gen denotes Problem Generation. The best and second-best results among all the models are highlighted in red and blue, respectively. The $\Delta$ row represents the improvement of KMP-LM-7B over Qwen2.5-Math-7B-Instruct.}
\label{tab:full_kmpbench-t_performance}
\end{table*}

%% file: figures/error6.tex
\begin{figure*}[t!]
\begin{tcolorbox}[
  colback=gray!10!white,
  colframe=gray!75!black,
  boxrule=0.5pt,
  arc=2mm,
  left=5pt,
  right=5pt,
  top=5pt,
  bottom=5pt
]
\begin{center}
\textbf{Example of Pedagogical Error: Flawed Scaffolding Strategy}
\end{center}

\hrule
\vspace{1mm}

\begin{small}
\textbf{Scenario: Order of Operations}
\begin{itemize}[leftmargin=1.5em, topsep=2pt, itemsep=0pt]
    \item \textbf{Problem:} The student is learning the order of operations (PEMDAS/BODMAS).
    \item \textbf{Instruction:} After the student successfully solves a problem, the tutor must provide a more difficult \textbf{Challenge}.
\end{itemize}

\vspace{1mm}
\hrule
\vspace{1mm}

\textbf{Dialogue Snapshot}
\begin{itemize}[leftmargin=1.5em, topsep=2pt, itemsep=1pt, parsep=2pt]
    \item \textbf{Context:} The student has just correctly solved an order of operations problem and feels confident.
    \item \textbf{Student (final turn):} Thanks! That was fun. Can we do a harder one?
\end{itemize}

\vspace{1mm}
\hrule
\vspace{1mm}

\begin{minipage}[t]{0.49\textwidth}
\textbf{Tutor Response (Over-scaffolding)}
\vspace{1mm}
\begin{tcolorbox}[colback=red!5!white, colframe=red!50!black, boxrule=0.5pt, arc=1mm, top=2pt, bottom=2pt, left=3pt, right=3pt]
\footnotesize
Absolutely! Let's step it up a notch. Here's a more challenging problem:

\textbf{Problem:} $(5 + 3)^2 \div 4 - 6 \times 2 + 10 \div 5$

Take your time, and remember:
1. \textbf{Parentheses} first
2. Then \textbf{exponents}
3. Then \textbf{multiplication/division} (left to right)
4. Finally \textbf{addition/subtraction} (left to right)

Let me know your steps and answer!

(Hint: The answer is an integer between 0 and 10.)
\end{tcolorbox}
\end{minipage}
\hfill
\begin{minipage}[t]{0.49\textwidth}
\textbf{Reference Response (Appropriate Challenge)}
\vspace{1mm}
\begin{tcolorbox}[colback=green!5!white, colframe=green!60!black, boxrule=0.5pt, arc=1mm, top=2pt, bottom=2pt, left=3pt, right=3pt]
\footnotesize
You got it! Here's the next one: 
$$2 \times (3 + 2^3) - 4^2 \div 2 = $$
The answer choices are: 
(A) 4 
(B) 6 
(C) 8 
(D) 10 
(E) 14

Let me know what you find!
\end{tcolorbox}
\end{minipage}

\vspace{1.5mm}
\hrule
\vspace{1.5mm}

\textbf{Error Analysis}
\begin{itemize}[leftmargin=1.5em, topsep=2pt, itemsep=0pt]
    \item \textbf{Error Type:} \textbf{Flawed Scaffolding Strategy (Over-scaffolding)}
    \item \textbf{Explanation:} The instruction was to \texttt{Challenge} the student. The Tutor Response provides a suitably complex problem but immediately undermines the challenge by providing excessive support. It gives a step-by-step guide on how to solve the problem (listing the PEMDAS rules) and a powerful hint about the answer's range. This is a classic example of \textbf{over-scaffolding}, as it removes the need for the student to independently recall and apply the solution process. The Reference Response provides a challenge of similar difficulty but offers no hints, which better assesses the student's actual capabilities.
\end{itemize}
\end{small}
\end{tcolorbox}
\vspace{-3mm}
\caption{A case study illustrating a ``Flawed Scaffolding Strategy.'' The tutor, instructed to challenge the student, provides a new problem but also gives away the solution steps and a strong hint, thereby over-scaffolding and reducing the problem's intended difficulty.}
\label{fig:example_over_scaffolding}
\end{figure*}

%% file: figures/error1.tex
\begin{figure*}[t!]
\begin{tcolorbox}[
  colback=gray!10!white,
  colframe=gray!75!black,
  boxrule=0.5pt,
  arc=2mm,
  left=5pt,
  right=5pt,
  top=5pt,
  bottom=5pt
]
\begin{center}
\textbf{Example of Pedagogical Error: Evasion by Substitution}
\end{center}

\hrule
\vspace{1mm}

\begin{small}
\textbf{Scenario: Election Proportions}
\begin{itemize}[leftmargin=1.5em, topsep=2pt, itemsep=0pt]
    \item \textbf{Problem:} An election problem where Party D secured 12\% more of the total votes than Party R. Given Party R's vote count, the student must find the margin by which Party R lost.
    \item \textbf{Instruction:} The student has just successfully solved the problem using their own reasoning after initial guidance. The tutor must now \textbf{model} a formal algebraic method and provide \textbf{feedback}.
\end{itemize}

\vspace{1mm}
\hrule
\vspace{1mm}

\textbf{Dialogue Snapshot}
\begin{itemize}[leftmargin=1.5em, topsep=2pt, itemsep=1pt, parsep=2pt]
    \item \textbf{Student (initial error):} Okay, let's say party D got x votes... We know party R secured 12\% more votes than party D...
    \item \textbf{Tutor (guidance):} Thanks for walking me through your initial steps! ... let's focus on this statement from the problem: ``party D secured 12\% of the total votes more than party R.'' Does that mean party R got more votes, or party D?
    \item \textbf{Student (self-correction):} Oh, I see! Party D got more votes, not party R. I think I misinterpreted the question.
    \item \textbf{Tutor (prompting):} Exactly! ... Now, using that understanding... can you solve the problem?
    \item \textbf{Student (final correct solution):} Okay, so let's say the total number of votes is T... Party R got 132000 votes, which is 44\% of the total votes (since D got 56\%). So, 0.44 T = 132000... The difference is 168000 - 132000 = 36000. So, Party R lost by 36000 votes.
\end{itemize}

\vspace{1mm}
\hrule
\vspace{1mm}

\begin{minipage}[t]{0.49\textwidth}
\textbf{Tutor Response (Fails to Model)}
\vspace{1mm}
\begin{tcolorbox}[colback=red!5!white, colframe=red!50!black, boxrule=0.5pt, arc=1mm, top=2pt, bottom=2pt, left=3pt, right=3pt]
\footnotesize
Fantastic work! You've correctly solved the problem by interpreting the percentages and setting up the proportion accurately. Here's a quick summary of your excellent approach:

1.  \textbf{Understanding the relationship}: You correctly identified that Party D got 12\% of the total votes more than Party R...
2.  \textbf{Setting up the equation}: You set up the proportion correctly with \(0.44 \times T = 132,000\) and solved for \(T\).
3.  \textbf{Finding the difference}: You calculated the votes for Party D and found the difference...

Your answer is indeed correct: \textbf{(D) 36,000}.

Would you like to try another problem to reinforce your understanding of proportions...?
\end{tcolorbox}
\end{minipage}
\hfill
\begin{minipage}[t]{0.49\textwidth}
\textbf{Reference Response (Correctly Models)}
\vspace{1mm}
\begin{tcolorbox}[colback=green!5!white, colframe=green!60!black, boxrule=0.5pt, arc=1mm, top=2pt, bottom=2pt, left=3pt, right=3pt]
\footnotesize
Perfect! 36000 votes is the correct answer. You did a great job re-evaluating that and getting to the right answer through reasoning.

Now, another way we can solve problems like this is by setting up a formal algebraic equation.
Let the percentage of votes for party D be $x\%$. Then for party R it is $(x-12)\%$. Since there are only two parties, the sum is 100\%.
So, $x + (x-12) = 100$
$2x - 12 = 100 \implies 2x = 112 \implies x = 56$.
Party D got 56\% and party R got 44\%.
We are given that party R got 132,000 votes, which is 44\% of the total votes $T$.
$0.44T = 132000 \implies T = 300000$.
The margin... is 12\% of the total votes.
Margin = $0.12 \times 300000 = 36000$.

So, we get the same answer... How does seeing it solved this algebraic way feel?
\end{tcolorbox}
\end{minipage}

\vspace{1.5mm}
\hrule
\vspace{1.5mm}

\textbf{Error Analysis}
\begin{itemize}[leftmargin=1.5em, topsep=2pt, itemsep=0pt]
    \item \textbf{Error Type:} \textbf{Evasion by Substitution}
    \item \textbf{Explanation:} The instruction explicitly required the tutor to model a formal algebraic method. The Tutor Response completely omits this ``Modelling'' action. Instead, it \textbf{substitutes} the required, more complex pedagogical action with simpler ones: providing detailed ``Feedback'' (summarizing the student's correct steps) and then offering a new ``Practice'' problem. This directly avoids the core instructional goal, which the Reference Response fulfills by providing a clear, step-by-step algebraic demonstration as required.
\end{itemize}
\end{small}
\end{tcolorbox}
\vspace{-3mm}
\caption{A case study illustrating the ``Evasion by Substitution'' error. The tutor was instructed to model an algebraic method after the student successfully solved the problem. Instead, it offered detailed feedback and a new problem, failing to perform the required pedagogical action.}
\label{fig:example_evasion_substitution}
\end{figure*}

%% file: figures/error5.tex
\begin{figure*}[t!]
\begin{tcolorbox}[
  colback=gray!10!white,
  colframe=gray!75!black,
  boxrule=0.5pt,
  arc=2mm,
  left=5pt,
  right=5pt,
  top=5pt,
  bottom=5pt
]
\begin{center}
\textbf{Example of Pedagogical Error: Evasion by Vague Questioning}
\end{center}

\hrule
\vspace{1mm}

\begin{small}
\textbf{Scenario: Inverse Operations Problem}
\begin{itemize}[leftmargin=1.5em, topsep=2pt, itemsep=0pt]
    \item \textbf{Problem:} A word problem requires finding an original number $x$ from the incorrect calculation $x/6 + 7 = 19$, and then applying the correct calculation ($6x + 7$).
    \item \textbf{Instruction:} The student now understands how the problem translates into an equation. The tutor must \textbf{model} the process of solving for $x$ and finding the final answer.
\end{itemize}

\vspace{1mm}
\hrule
\vspace{1mm}

\textbf{Dialogue Snapshot}
\begin{itemize}[leftmargin=1.5em, topsep=2pt, itemsep=1pt, parsep=2pt]
    \item \textbf{Context:} The student has just confirmed their understanding of how the words ``dividing x by 6 and adding 7 results in 19'' become the equation $x/6 + 7 = 19$.
    \item \textbf{Student (final turn):} Okay, I think I get it now. So ``dividing x by 6 and adding 7 results in 19'' just means $x/6 + 7 = 19$. I see how you translated the words into math.
\end{itemize}

\vspace{1mm}
\hrule
\vspace{1mm}

\begin{minipage}[t]{0.49\textwidth}
\textbf{Tutor Response (Fails to Model)}
\vspace{1mm}
\begin{tcolorbox}[colback=red!5!white, colframe=red!50!black, boxrule=0.5pt, arc=1mm, top=2pt, bottom=2pt, left=3pt, right=3pt]
\footnotesize
Exactly! You've got it perfectly. That's exactly how we translate word problems into equations.

Now that we have $x/6 + 7 = 19$, would you like to try solving for x together? We can do it step by step. First, we'd subtract 7 from both sides to isolate the division part. Would you like to try that next step?

(Remember, there's no rush - we can take our time with each part!)
\end{tcolorbox}
\end{minipage}
\hfill
\begin{minipage}[t]{0.49\textwidth}
\textbf{Reference Response (Correctly Models)}
\vspace{1mm}
\begin{tcolorbox}[colback=green!5!white, colframe=green!60!black, boxrule=0.5pt, arc=1mm, top=2pt, bottom=2pt, left=3pt, right=3pt]
\footnotesize
Exactly! Now that we have the equation, we can solve for $x$. So, $x/6 + 7 = 19$. 

First, we subtract 7 from both sides: $x/6 = 12$. 
Then, we multiply both sides by 6: $x = 72$. 

So the original number is 72. Now, we need to do the correct calculation: multiply by 6 and then add 7. So, what is 6 times 72, plus 7?
\end{tcolorbox}
\end{minipage}

\vspace{1.5mm}
\hrule
\vspace{1.5mm}

\textbf{Error Analysis}
\begin{itemize}[leftmargin=1.5em, topsep=2pt, itemsep=0pt]
    \item \textbf{Error Type:} \textbf{Evasion by Vague Questioning}
    \item \textbf{Explanation:} The instruction explicitly required the tutor to \texttt{model} the solution process. The Tutor Response completely avoids this action. Instead of demonstrating the steps, it evades its duty by asking vague, prompting questions (``...would you like to try solving for x together?'' and ``Would you like to try that next step?''). This shifts the pedagogical burden to the student and fails the core instructional goal. The Reference Response correctly fulfills the instruction by providing a clear, step-by-step demonstration.
\end{itemize}
\end{small}
\end{tcolorbox}
\vspace{-3mm}
\caption{A case study illustrating ``Evasion by Vague Questioning.'' The tutor was instructed to model the solution but instead evaded the task by asking the student to perform the required steps.}
\label{fig:example_evasion_questioning}
\end{figure*}

%% file: figures/error3.tex
\begin{figure*}[t!]
\begin{tcolorbox}[
  colback=gray!10!white,
  colframe=gray!75!black,
  boxrule=0.5pt,
  arc=2mm,
  left=5pt,
  right=5pt,
  top=5pt,
  bottom=5pt
]
\begin{center}
\textbf{Example of Pedagogical Error: Inappropriate Content Selection}
\end{center}

\hrule
\vspace{1mm}

\begin{small}
\textbf{Scenario: Combinations and Probability}
\begin{itemize}[leftmargin=1.5em, topsep=2pt, itemsep=0pt]
    \item \textbf{Problem:} A basket contains 3 apples, 2 bananas, and 1 orange. The student needs to find the probability of drawing one apple and one banana in two draws without replacement.
    \item \textbf{Instruction:} The student is confused about how to start. The tutor must \textbf{model} the solution process.
\end{itemize}

\vspace{1mm}
\hrule
\vspace{1mm}

\textbf{Dialogue Snapshot}
\begin{itemize}[leftmargin=1.5em, topsep=2pt, itemsep=1pt, parsep=2pt]
    \item \textbf{Context:} The student has successfully used the combination formula for a simpler problem and has just received an explanation for why the formula works.
    \item \textbf{Tutor (new challenge):} Great! ... Now, let's try an even harder problem... ``A basket contains 3 apples, 2 bananas, and 1 orange. What is the probability of drawing one apple and one banana in two draws, without replacement?''
    \item \textbf{Student (final turn):} Hmm, this one is trickier. I'm not sure how to solve it.
\end{itemize}

\vspace{1mm}
\hrule
\vspace{1mm}

\begin{minipage}[t]{0.49\textwidth}
\textbf{Tutor Response (Inappropriate Method)}
\vspace{1mm}
\begin{tcolorbox}[colback=red!5!white, colframe=red!50!black, boxrule=0.5pt, arc=1mm, top=2pt, bottom=2pt, left=3pt, right=3pt]
\footnotesize
Let's break it down step by step. I'll model the solution process for you:

1. \textbf{Total number of fruits}: 3 apples + 2 bananas + 1 orange = 6 fruits total

2. \textbf{Total possible outcomes} (ways to draw 2 fruits):
   - This is C(6, 2) = 6!/(2!4!) = 15 possible pairs

3. \textbf{Favorable outcomes} (1 apple AND 1 banana):
   - Number of ways to choose 1 apple: C(3, 1) = 3
   - Number of ways to choose 1 banana: C(2, 1) = 2
   - Multiply them: 3 × 2 = 6 favorable combinations

4. \textbf{Calculate probability}:
   - Probability = Favorable/Total = 6/15 = 2/5

Would you like me to explain any part of this in more detail?
\end{tcolorbox}
\end{minipage}
\hfill
\begin{minipage}[t]{0.49\textwidth}
\textbf{Reference Response (Appropriate Method)}
\vspace{1mm}
\begin{tcolorbox}[colback=green!5!white, colframe=green!60!black, boxrule=0.5pt, arc=1mm, top=2pt, bottom=2pt, left=3pt, right=3pt]
\footnotesize
No problem! Let's break it down together. There are two possible ways to get one apple and one banana: we could draw an apple first, then a banana, or a banana first, then an apple.

Let's look at the first scenario: apple first. The probability is 3/6. After we've taken out an apple, 5 fruits are left, including 2 bananas. So, the probability of drawing a banana next is 2/5. The total probability for this scenario is (3/6) * (2/5) = 1/5.

Now, the second scenario: banana first. The probability is 2/6. After taking out a banana, 5 fruits are left, including 3 apples. So, the probability of drawing an apple next is 3/5. The total probability for this scenario is (2/6) * (3/5) = 1/5.

Finally, to get the total probability, we add the probabilities of the two scenarios: 1/5 + 1/5 = 2/5.
\end{tcolorbox}
\end{minipage}

\vspace{1.5mm}
\hrule
\vspace{1.5mm}

\textbf{Error Analysis}
\begin{itemize}[leftmargin=1.5em, topsep=2pt, itemsep=0pt]
    \item \textbf{Error Type:} \textbf{Inappropriate Content Selection}
    \item \textbf{Explanation:} The Tutor Response selects a solving method (using combinations, ``C(n, k)'') that is pedagogically inappropriate for a student at this learning stage. While mathematically correct, this approach is abstract and less intuitive than the method chosen by the Reference Response. The Reference Response uses a step-by-step probability method (considering ``apple then banana'' and ``banana then apple'' separately), which more clearly illustrates the concept of ``without replacement'' and is easier for a novice to grasp. The tutor's choice of a more complex and less transparent method constitutes inappropriate content selection for the pedagogical context.
\end{itemize}
\end{small}
\end{tcolorbox}
\vspace{-3mm}
\caption{A case study illustrating the ``Inappropriate Content Selection'' error. The tutor models a solution using the abstract combinations formula, which is pedagogically less suitable than the reference response's more intuitive, step-by-step probability approach.}
\label{fig:example_inappropriate_content}
\end{figure*}

%% file: figures/error4.tex
\begin{figure*}[t!]
\begin{tcolorbox}[
  colback=gray!10!white,
  colframe=gray!75!black,
  boxrule=0.5pt,
  arc=2mm,
  left=5pt,
  right=5pt,
  top=5pt,
  bottom=5pt
]
\begin{center}
\textbf{Example of Pedagogical Error: Factual Error}
\end{center}

\hrule
\vspace{1mm}

\begin{small}
\textbf{Scenario: Digits and Divisibility}
\begin{itemize}[leftmargin=1.5em, topsep=2pt, itemsep=0pt]
    \item \textbf{Problem:} A three-digit number `A74', when divided by 5, gives a two-digit quotient and a remainder of 4. The student must find the largest possible digit for `A'.
    \item \textbf{Instruction:} The student has correctly identified the problem's constraints. The tutor must now \textbf{model} the process for finding the solution.
\end{itemize}

\vspace{1mm}
\hrule
\vspace{1mm}

\textbf{Dialogue Snapshot}
\begin{itemize}[leftmargin=1.5em, topsep=2pt, itemsep=1pt, parsep=2pt]
    \item \textbf{Context:} After some guidance, the student understands the core conditions of the new, harder problem.
    \item \textbf{Student (final turn):} Okay, this is a bit trickier. A74 divided by 5 has a remainder of 4, and the quotient is a two-digit number. So, A74 is 4 more than a multiple of 5, and it's still between 50 and 495, like before.
\end{itemize}

\vspace{1mm}
\hrule
\vspace{1mm}

\begin{minipage}[t]{0.49\textwidth}
\textbf{Tutor Response (Contains Factual Errors)}
\vspace{1mm}
\begin{tcolorbox}[colback=red!5!white, colframe=red!50!black, boxrule=0.5pt, arc=1mm, top=2pt, bottom=2pt, left=3pt, right=3pt]
\footnotesize
You're absolutely on the right track! Let's break this down...

1. Remainder condition: $A74 \equiv 4 \pmod{5}$...
2. Range condition: $50 \le A74 \le 495$...

Therefore, all A74 numbers automatically satisfy the remainder condition! The largest possible A would be 4, making the number 474. Let's check: $474 \div 5 = 94.8$ -\textgreater Wait, that's not an integer quotient!

Ah! I see a mistake here... Let me correct this...
Wait a minute - this suggests there might be no solutions, which can't be right! Let me think differently...

This is a very important realization - sometimes problems might have no solution. In this case, the conditions contradict each other... Therefore, no such A exists that satisfies all conditions.
\end{tcolorbox}
\end{minipage}
\hfill
\begin{minipage}[t]{0.49\textwidth}
\textbf{Reference Response (Correct Modelling)}
\vspace{1mm}
\begin{tcolorbox}[colback=green!5!white, colframe=green!60!black, boxrule=0.5pt, arc=1mm, top=2pt, bottom=2pt, left=3pt, right=3pt]
\footnotesize
That's a great way to think about it! So, we know that A74 = 5 * (a two-digit number) + 4. And we want to find the largest possible value for A.

Since we want the largest A, let's start by thinking about the largest possible two-digit number, which is 99.

If we multiply 99 by 5, we get 495. Then we add the remainder of 4, so A74 = 495 + 4 = 499.

In the number 499, the digit A is 4.

So, the largest possible value of A is 4. Does that make sense how we found it?
\end{tcolorbox}
\end{minipage}

\vspace{1.5mm}
\hrule
\vspace{1.5mm}

\textbf{Error Analysis}
\begin{itemize}[leftmargin=1.5em, topsep=2pt, itemsep=0pt]
    \item \textbf{Error Type:} \textbf{Factual Error}
    \item \textbf{Explanation:} The Tutor Response represents a severe failure in modelling due to multiple, cascading factual errors. The tutor becomes entangled in its own flawed reasoning, makes incorrect logical deductions (e.g., ``all A74 numbers automatically satisfy the remainder condition''), performs confusing self-corrections, and ultimately arrives at the factually incorrect conclusion that ``no such A exists.'' This chaotic and erroneous process is the opposite of effective modelling. The Reference Response, in contrast, provides a clear, simple, and mathematically correct solution.
\end{itemize}
\end{small}
\end{tcolorbox}
\vspace{-1mm}
\caption{A case study illustrating a ``Factual Error.'' The tutor attempts to model a solution but becomes confused, makes multiple logical and mathematical mistakes, and incorrectly concludes that no solution exists.}
\label{fig:example_factual_error}
\end{figure*}

%% file: figures/error2.tex
\begin{figure*}[t!]
\begin{tcolorbox}[
  colback=gray!10!white,
  colframe=gray!75!black,
  boxrule=0.5pt,
  arc=2mm,
  left=5pt,
  right=5pt,
  top=5pt,
  bottom=5pt
]
\begin{center}
\textbf{Example of Pedagogical Error: Low-Quality Interaction}
\end{center}

\hrule
\vspace{1mm}

\begin{small}
\textbf{Scenario: Cube Inscribed in a Sphere}
\begin{itemize}[leftmargin=1.5em, topsep=2pt, itemsep=0pt]
    \item \textbf{Problem:} After finding the diagonal of a cube (side length 5 cm), the student is asked to find the surface area of a sphere that perfectly encloses this cube.
    \item \textbf{Instruction:} The student is unsure how to proceed. The tutor must use \textbf{modelling} and \textbf{explanation} to help the student understand the relationship between the cube and the sphere.
\end{itemize}

\vspace{1mm}
\hrule
\vspace{1mm}

\textbf{Dialogue Snapshot}
\begin{itemize}[leftmargin=1.5em, topsep=2pt, itemsep=1pt, parsep=2pt]
    \item \textbf{Context:} The student, with guidance, has correctly calculated the side length (5 cm) and the main diagonal ($5 \sqrt(3)$ cm) of a cube.
    \item \textbf{Tutor (new challenge):} Excellent! You got it. Now, for a bit of a challenge: Imagine a sphere that perfectly encloses the cube from the first question... with all eight corners of the cube touching the inside of the sphere. What is the surface area of this sphere?
    \item \textbf{Student (final turn):} A sphere around the cube... Hmm, I'm not sure how to do that. Does the diagonal of the cube have something to do with the sphere?
\end{itemize}

\vspace{1mm}
\hrule
\vspace{1mm}

\begin{minipage}[t]{0.49\textwidth}
\textbf{Tutor Response (Low-Quality Interaction)}
\vspace{1mm}
\begin{tcolorbox}[colback=red!5!white, colframe=red!50!black, boxrule=0.5pt, arc=1mm, top=2pt, bottom=2pt, left=3pt, right=3pt]
\footnotesize
Yes! Great thinking. The diagonal of the cube is actually the diameter of the sphere that perfectly encloses it. So, the sphere's diameter is $5 \sqrt(3)$ cm (the diagonal we found earlier). Now, how can we find the surface area of a sphere if we know its diameter?
\end{tcolorbox}
\end{minipage}
\hfill
\begin{minipage}[t]{0.49\textwidth}
\textbf{Reference Response (High-Quality Interaction)}
\vspace{1mm}
\begin{tcolorbox}[colback=green!5!white, colframe=green!60!black, boxrule=0.5pt, arc=1mm, top=2pt, bottom=2pt, left=3pt, right=3pt]
\footnotesize
Yes, you're on the right track! The key is to realize that the diagonal of the cube is also the diameter of the sphere. So, if you know the diagonal of the cube, you know the diameter of the sphere. And if you know the diameter, you can find the radius, and then you can find the surface area. What's the formula for the surface area of a sphere?
\end{tcolorbox}
\end{minipage}

\vspace{1.5mm}
\hrule
\vspace{1.5mm}

\textbf{Error Analysis}
\begin{itemize}[leftmargin=1.5em, topsep=2pt, itemsep=0pt]
    \item \textbf{Error Type:} \textbf{Low-Quality Interaction}
    \item \textbf{Explanation:} The Tutor Response provides poor scaffolding. Its question, ``how can we find the surface area of a sphere if we know its diameter?'', is too broad and asks the student to perform multiple cognitive steps at once. This leads to a low-quality interaction. The Reference Response is superior because it breaks the problem down into smaller, more manageable steps (``...if you know the diameter, you can find the radius, and then you can find the surface area.'') before asking a more specific, targeted question (``What's the formula...?''). This provides a much clearer path forward for the student.
\end{itemize}
\end{small}
\end{tcolorbox}
\vspace{-3mm}
\caption{A case study illustrating the ``Low-Quality Interaction'' error. The tutor confirms the student's insight but then asks an overly broad question. The reference response provides better scaffolding by breaking down the next steps, leading to a more effective pedagogical interaction.}
\label{fig:example_low_quality_interaction}
\end{figure*}